\documentclass[twocolumn, 3p]{elsarticle}

\usepackage{lineno}
\usepackage{hyperref}

\usepackage{color} 
\usepackage{rotating}

\usepackage{multirow}
\usepackage{amssymb}
\usepackage{listings}
\usepackage{multirow}
\usepackage{slashbox}
\usepackage{array}
\usepackage{enumerate}
\usepackage{amsmath}
\usepackage{bm}
\usepackage{hyperref}
\usepackage{stfloats}

\usepackage{graphicx}
\usepackage{algorithmic}
\usepackage{algorithm}
\usepackage{subfigure}

\usepackage{setspace}
\usepackage{booktabs}

\usepackage{lscape}	

\usepackage{placeins}	

\begin{document}

\begin{frontmatter}

\title{APPD: Adaptive and Precise Pupil Boundary Detection\\using Entropy of Contour Gradients\tnoteref{t1}}

\tnotetext[t1]{This work is supported by The Scientific and Technological Research Council of Turkey (TUBITAK) and Anadolu University Commission of Scientific Research Projects (BAP) under the grant numbers 111E053 and 1207F113, respectively.}

\tnotetext[t2]{Cuneyt Akinlar was with Department of Computer Engineering, Anadolu University during the time of this study.}

%
%
%

\author[auEemAddress]{Cihan Topal}
\ead{cihant@anadolu.edu.tr}

\author[auCeAddress]{Halil Ibrahim Cakir}
\ead{halilibrahimcakir@gmail.com}

\author[auCeAddress]{Cuneyt Akinlar}
\ead{cuneytakinlar@gmail.com}

\address[auEemAddress]{Department of Electrical and Electronics Engineering, Anadolu University, 26470 Eskisehir, Turkey}


\address[auCeAddress]{Department of Computer Engineering, Anadolu University, 26470 Eskisehir, Turkey\tnoteref{t2}}

\begin{abstract}

Eye tracking spreads through a vast area of applications from ophthalmology, assistive technologies to gaming and virtual reality.
Precisely detecting the pupil's contour and center is the very first step in many of these tasks, hence needs to be performed accurately.
Although detection of pupil is a simple problem when it is entirely visible; occlusions and oblique view angles complicate the solution. 
In this study, we propose APPD, an adaptive and precise pupil boundary detection method that is able to infer whether entire pupil is in clearly visible by a heuristic that estimates the shape of a contour in a computationally efficient way.
Thus, a faster detection is performed with the assumption of no occlusions.
If the heuristic fails, a more comprehensive search among extracted image features is executed to maintain accuracy.
Furthermore, the algorithm can find out if there is no pupil as an helpful information for many applications.
We provide a dataset containing 3904 high resolution eye images collected from 12 subjects and perform an extensive set of experiments to obtain quantitative results in terms of accuracy, localization and timing.
The proposed method outperforms three other state of the art algorithms and has an average execution time $\sim$5 ms in single-thread on a standard laptop computer for 720p images. 

\end{abstract}


\end{frontmatter}


\section{Introduction}
\label{sec:intro}

Eye tracking (ET) has emerged as an important research area with a diverse set of applications including human computer interaction~\cite{Galdi2016, Cheng2017, Ko2008}, diagnosis of psychological, neurological and ophthalmologic individuals~\cite{Hong2017, Nystrom2017}, assistive systems for drivers and disabled people~\cite{Soltani2016, Cristina2016, Topal2008}, marketing research~\cite{Tzafilkou2017, Mou2018}, and biometrics~\cite{Kumar2010, Sahmoud2013, Raffei2013, Abate2015}. 
Further details on application areas of eye tracking can be found in~\cite{Duchowski2002}. 

Pupil boundary detection and center estimation are essential steps in all eye tracking systems and have to be performed precisely.
In point-of-gaze (PoG) detection, extraction of pupil center is required to estimate the location of gaze. 
In such applications, loss of even a single pixel precision in the pupil center may cause a noticeable error in the gaze direction vector which can cause a significant drift in the estimated gaze.
Pupil boundary detection is also an important step that has to be performed accurately for biometric applications and medical studies.
Iris recognition is the most in demand biometric application that starts with the extraction of pupil boundary lying in between the pupil and the iris. 

Another emerging application area of eye tracking is virtual reality (VR) technologies which recently have a significant leap in popularity.
There are efforts to integrate ET technology in VR studies to increase the feeling of immersion via rendering virtual environment with a depth of field effect similar to human vision.
VR technology renders a 3D scene from two different point of views, i.e. from the views of left and right eyes of a user.
To prevent problems such as motion sickness, 3D locations of rendered objects should be aligned with respect to the interpupillary distance of the user.
Moreover, developers integrate eye tracking into VR systems to better simulate the human visual system.
They sharply render image locations where the user focuses on and blur other image regions with respect to depth information to enhance the immersion effect.
Therefore, accurate and fast detection of pupil paves the way for a better VR experience.

In this study, we propose APPD, a novel \textit{A}daptive and \textit{P}recise \textit{P}upil boundary \textit{D}etection algorithm which takes an eye image and works by extracting arcs from the edge segments of the image and joining them to find the pupil boundary, hence the center.
Organization of the paper is as follows; 
we give a comprehensive related work on pupil boundary and center detection in Section~\ref{sec:RelatedWork}, 
we explain our method in more detail in Section~\ref{sec:ProposedMethod}, 
quantitative and qualitative assessment of the proposed method with three state of the art algorithms and analysis of running times are presented in Section~\ref{sec:ExperimentalResults}, 
and we finalize the paper with concluding remarks in Section~\ref{sec:Conclusions}.

\section{Related Work}
\label{sec:RelatedWork}

The literature on pupil detection is very rich due to the diversity of applications and variety of approaches have been proposed.
In this section, our goal is to give a high-level picture of the proposed solutions for pupil boundary detection and/or pupil center estimation.

Many early methods in the literature utilize discriminative visual structure of human eye to detect the pupil.
Dark intensity of pupil region and its high contrast compare to bright sclera region offers a relatively easy way to solve the problem.
In this manner, many algorithms extracts pupil center (or limbus center in iris recognition studies) via combinations of several methods like thresholding, morphological operations, connected component analysis and center of mass algorithms with various additional steps~\cite{Goni2004, Maenpaa2005, Long2007, Keil2010, Lin2010}.
In addition, there are methods which also benefit model fitting approaches to find pupil or iris boundary as a circle or ellipse~\cite{Wang2002, Ma2004, Dey2007, Agustin2010}. 
In these studies, edge and contour extraction is employed and followed by Hough transform or RANSAC algorithm to accurately estimate the boundary~\cite{Ballard1981, Fischler1981}.
Hough transform is a model-based technique which aims to detect geometric primitives by a voting scheme~\cite{Ballard1981}.
Since it is a model based method, it is robust against occlusions, however, it may give many false detections as well for the same reason.
RANSAC is an iterative method to estimate the parameters of a mathematical or geometric model from a set of samples with outliers~\cite{Fischler1981}.
At each iteration, it randomly samples among the data and eliminates the outliers by checking their relevance to the model.

Dark/bright pupil differencing is another common approach to roughly detect eye locations in a remotely taken image~\cite{Ebisawa1998, Hiley2006, Morimoto2000}.
It works by differencing two successive frames that captured with on-axis and off-axis illumination, respectively.
Due to the physical structure of human eye, on-axis illumination causes a significant brightness inside the pupil.
Therefore, pupil regions become more salient in the difference image.

Along with the feature-based methods, there are also purely model-based approaches which are mostly utilized in iris recognition studies.
Daugman~\cite{Daugman2002} proposes an integro-differential operator for detecting the pupil and iris boundaries aiming to maximize the contour integral value on the smoothed image derivative: 
%
\begin{equation}
	\label{eq:Eq_2_1} 
	{\max}_{(r,x_0,y_0)} \left|G_\sigma(r)*\frac{\partial}{\partial r}\oint_{(r,x_0,y_0)}\frac{I(x,y)}{2\pi r} ds \right|
\end{equation}
where $I(x,y)$ is the 2D array of intensity values in the image; $r, x_0, y_0$ denote the radius and center coordinates for various circular regions; $G_\sigma(r)$ is the employed Gaussian filter with a standard deviation $\sigma$; and $s$ is the contour of the circle given by $(r, x_0, y_0)$.

Arvacheh and Tizhoosh~\cite{Arvacheh2006} developed an iterative algorithm based on an active counter model which is also capable of detecting near-circular shapes.
These methods work fine for the shapes closer to a circle rather than an ellipse, however, they require a model-based search in the image plane to find $r,x_{0},y_{0}$ parameters that maximize the response to the given model.
Since model-based search is computationally expensive and robust to only a narrow pose range; it cannot be employed in real-time eye tracking applications.

Zhu et al.~\cite{Zhu1999} use curvature of pupil contour to sort out boundary pixels which belong to prospective occlusions.
They detect blobs in the binarized image and then extract the contour of the biggest blob.
Finally, edge pixels of the pupil boundary are selected by employing a set of heuristics (e.g. eyelids have positive curvature, etc.) and ellipse fit applied to chosen pixels.

Another interesting approach on pupil detection is proposed and utilized in EyeSeeCam project~\cite{Kumar2009}.
The algorithm extracts edge segments, then removes glints (reflections on the cornea surface originated from the light sources of the eye tracker) and other unfavourable artefacts by a sequence of morphological operations based on several assumptions.
Finally, Delaunay Triangulation is applied to remaining pixels and pupil boundary is detected assuming it is a convex hull.

\begin{table*}[!t]
	\begin{onehalfspace}
		
		\caption{A brief taxonomy for pupil boundary detection / center estimation algorithms.}
		\scalebox{0.8} 
		{
			\begin{tabular}{lcccccccccc}
				\hline			
				
				{\textbf{{\large Algorithms}}} 
				& \begin{sideways} \parbox{25mm} {\centering Downscaling } \end{sideways}
				
				& \begin{sideways} \parbox{25mm} {\begin{singlespace} \centering  Bright/Dark Pupil Differencing  \end{singlespace}} \end{sideways}			
				
				& \begin{sideways} \parbox{25mm} {\begin{singlespace} \centering Image Thresholding / Binarization \end{singlespace}} \end{sideways} 
				
				& \begin{sideways} \parbox{23mm} {\begin{singlespace} \centering Morphological Operations \end{singlespace} } \end{sideways} 
				
				& \begin{sideways} \parbox{25mm}{\centering Edge Detection}                   \end{sideways} 
				
				& \begin{sideways} \parbox{27mm}{\begin{singlespace} \centering Blob Detection/ Connected Comp. Analysis \end{singlespace}} \end{sideways} 
				
				& \begin{sideways} \parbox{23mm}{\begin{singlespace} \centering Circle / Ellipse Fitting \end{singlespace}} \end{sideways} 
				
				& \begin{sideways} \parbox{23mm}{\begin{singlespace} \centering Center of Mass Algorithm \end{singlespace} } \end{sideways} 
				
				& \begin{sideways} \parbox{23mm}{\begin{singlespace} \centering Use of Temporal Information \end{singlespace} } \end{sideways} 
				
				& \begin{sideways} \parbox{23mm}{ \begin{singlespace} \centering Other / Comments  \end{singlespace} } \end{sideways}  \\  \hline

				
				G\~oni et al.~\cite{Goni2004} &  &  & adaptive & \textbullet &  &  &  & modified & \textbullet  & \small{$\varnothing$} \\ \hline
				
				M\"aenp\"a\"a~\cite{Maenpaa2005} &  &  & \textbullet & \textbullet &  &  &  & \textbullet &  & 	$\triangle$ \\ \hline
				
				Long et al.~\cite{Long2007} &  &  & \textbullet &  &  & \textbullet &  & symmetric &  & $\square$, \small{$\varnothing$} \\ \hline
				
				Keil et al.~\cite{Keil2010} &  &  & \textbullet &  & for glint &  &  & \textbullet &  & $\square$, \small{$\varnothing$}, $^1$ \\ \hline
				
				Lin et al.~\cite{Lin2010} & \textbullet &  & \textbullet & \textbullet & \textbullet &  &  & parallelogram &  & $\square$, \small{$\varnothing$} \\ \hline
				
				Wang and Sun~\cite{Wang2002} &  &  & \textbullet & \textbullet & vertical &  & ellipse &  &  & 	$\triangle$ \\ \hline
				
				Ma et al.~\cite{Ma2004} &  &  & adaptive &  & \textbullet &  & Hough\cite{Ballard1981} &  &  &  \\ \hline
				
				Dey and Samanta~\cite{Dey2007} & \textbullet &  &  &  & \textbullet & for edges & circle &  &  & $^1$ \\ \hline
				
				Agustin et al.~\cite{Agustin2010} &  &  & \textbullet &  &  &  & ellipse &  &  &  \\ \hline
				
				Ebisawa~\cite{Ebisawa1998} &  & \textbullet & \textbullet & \textbullet &  &  &  & \textbullet &  & \small{$\varnothing$} \\ \hline
				
				Hiley et al.~\cite{Hiley2006} &  & \textbullet & \textbullet & iterative &  &  &  & \textbullet &  & \small{$\varnothing$} \\ \hline
				
				Morimoto et al.~\cite{Morimoto2000} &  & \textbullet & \textbullet &  &  &  &  &  & \textbullet  & \small{$\varnothing$} \\ \hline
				
				Zhu et al.~\cite{Zhu1999} &  &  & \textbullet &  &  & \textbullet & ellipse &  &  & $^2$ \\ \hline
				
				Kumar et al.~\cite{Kumar2009} &  &  &  & \textbullet & \textbullet & for edges &  &  &  & $\square, ^3$ \\ \hline
				
				Li et al.~\cite{Li2005} &  &  &  &  & radial &  & ellipse &  &  & $^4$ \\ \hline
				
				\'{S}wirski et al.~\cite{Swirski2012} & & & k-means & \textbullet & \textbullet & & iterative & & & $\square, ^5$ \\ \hline
				
				Fuhl et al~\cite{Fuhl2016}  & \textbullet & & & \textbullet & \textbullet &  & ellipse & & & $^6$ \\ \hline
				
			\end{tabular}
		}
		
		\begin{onehalfspace}
			\scalebox{0.72} 
			{ 
				\begin{tabular}{cl}			
					
					$\varnothing$ & \hspace{-4mm} does not detect the pupil boundary, only estimates its center. \hspace{2mm} 
					$\triangle$  performs iris detection. \hspace{2mm}
					$\square$ performs ROI detection. \\
					
					$^1$ & \hspace{-5mm} applies histogram back-projection or non-linear power transform on the image to make the pupil more salient.  \\
					$^2$ & \hspace{-5mm} before ellipse fitting, tries to determine the false pupil contour pixels w.r.t. their curvature values by a set of heuristics. \\
					$^4$ & \hspace{-5mm} requires removal of glints. Assumes that the initial point for ray casting is inside the pupil. Iterative algorithm. \\
					$^3$ & \hspace{-5mm} performs Fast Radial Symmetry detection and Delaunay Triangulation. Removes glints and artefacts by a set of morphological assumptions. \\
					$^5$ & \hspace{-5mm} tries to find an ellipse that matches the edge image points and is orthogonal to the gradients of the image. \\
					$^6$ & \hspace{-5mm} detects and filter edges. Uses two different approaches, i.e. algorithmic and morphological. Rescales image if it fails in the first attempt.
				\end{tabular}
			}
		\end{onehalfspace}	
		\label{Taxonomy}
	\end{onehalfspace}
\end{table*}

Starburst algorithm~\cite{Li2005} 
estimates the pupil center by an iterative radial feature detection technique instead of finding all edges.
It starts by locating and removing glints if any exists.
Then, rays are cast from an initial point within a 20$^\circ$ of radial step.
Each ray stops where image derivative is greater than a threshold value, i.e., when a sharp intensity change occurs.
This operation is iterated with an updated starting point and a set of feature points are collected at each step.
Finally, ellipse fit is applied to the collected points with RANSAC~\cite{Fischler1981}.
In~\cite{Ryan2008} Ryan et al. aim to adapt the Starburst algorithm to elliptical iris segmentation problem.

\'{S}wirski et al.~\cite{Swirski2012} approximately detects pupil region by a Haar-like feature~\cite{Viola2001, Fischler1981}. 
Next, they apply \textit{k}-means segmentation to determine a proper pupil threshold.
Then a modified RANSAC-based ellipse fitting method is employed which utilizes gradient information as well as the spatial coordinates to find the pupil boundary.

In a more recent study Fuhl et al. detect edges on the eye image and filter them with respect to several morphological criteria~\cite{Fuhl2016}.
Later, edge segments are constructed from the remaining edge pixels and some of the edge segments (i.e. straight lines) are eliminated by various heuristics. 
Finally, remaining contours are tested by ellipse fitting and the best ellipse is selected by a cost function which utilize inner gray value and shape roundness. 

Table~\ref{Taxonomy} gives a brief taxonomy of the above-mentioned pupil detection algorithms.
As seen in the table, thresholding is a common technique in the literature. 
Despite thresholding can quickly discriminate high contrast image regions, it is highly vulnerable to lighting conditions and parameter configuration.
Consequently, it fails finding the exact location where the actual intensity change occurs and can decrease the localization accuracy.
Another frequently employed technique is morphological operations which are applied on the thresholded binary image to suppress remaining undesired pixel sets and improve modal structure of the image.
However, morphological operations may also degrade the actual information on the image and can cause significant errors on the result.
Similarly, algorithms that utilize thresholding and blob detection to find a center point for pupil are obviously not capable of detecting the boundary.
Hence, they cannot be applied on most biometrics or medical studies which requires precise detection of the boundaries of pupil and iris. 

%
\begin{figure*}[!t]
	\centerline{\includegraphics[width=1.03\linewidth]{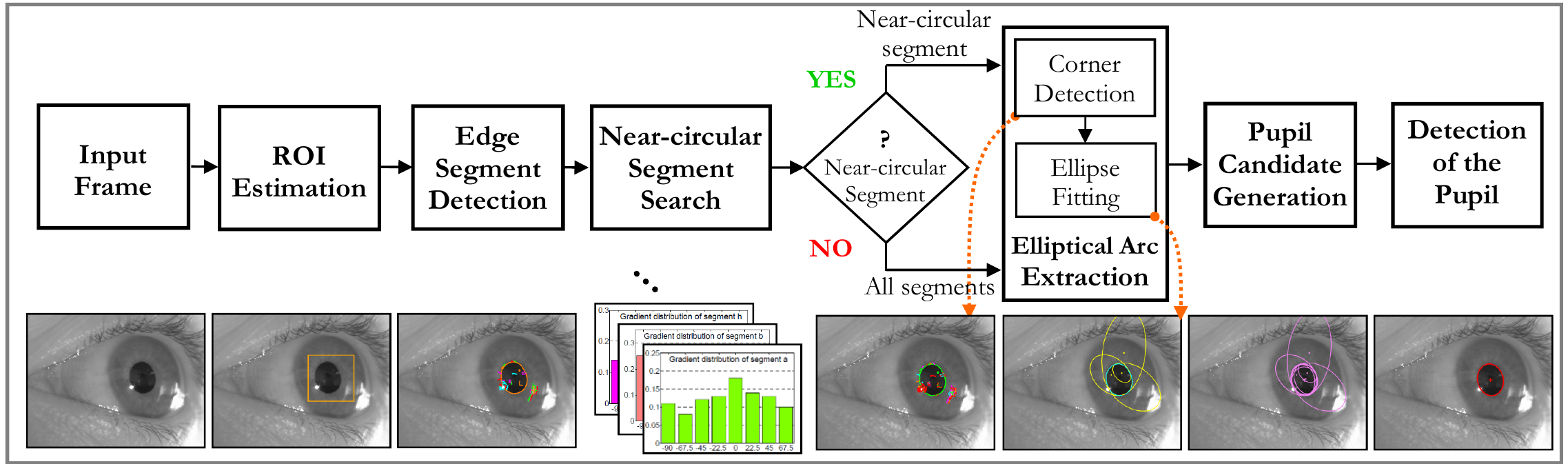}}
	\caption{Processing pipeline of the proposed algorithm.}
	\label{fig:SystemSchema}
\end{figure*}

Downscaling the image to save computational time has an obvious cost as it decreases the accuracy by wasting spatial resolution. 
Bright/dark pupil differencing requires a little amount of computation and eases roughly locating the pupil, however it has important drawbacks.
First, it needs additional hardware to obtain bright and dark pupil images consecutively in a synchronous manner.
Furthermore, it reduces temporal resolution since it needs two frames to perform a single detection.
Due to the same reason it is very sensitive to motion and it fails if a large displacement on pupil location occurs in between two consecutive frames.
Ebisawa specifically addresses this problem in~\cite{Ebisawa2009} and proposes various methods for positional compensation.

This section presents an overview of related studies covering both biometrics and eye tracking areas from the viewpoint of pupil detection problem.
For interested readers, there are also comprehensive surveys that review the gaze estimation literature; in particular~\cite{Hansen2010,Morimoto2005}.

In summary, according to the scientific and industrial developments, a pupil boundary detection algorithm has to be both precise and fast in order to pave the way for the next generation eye tracking applications.
There are studies in the literature that improve speed by methods such as downscaling and binarization, however, they significantly reduce the precision.
There are also studies providing accurate results but the computational burden they require hinders their applicability to real-time applications.
To address the gap for a fast and precise pupil detection algorithm, we propose APPD which both accurately estimates the pupil boundary and runs in a few milliseconds for high resolution images.
We explain steps of the proposed algorithm in the next section.

\section{Proposed Method}
\label{sec:ProposedMethod}

In this study, we propose APPD, an adaptive method for pupil boundary detection which is able to save computation by inferring whether an occlusion is the case or not.
Thus, the computation takes very little time if the pupil is in entirely visible by the camera.
On the contrary, algorithm infers if the pupil is severely occluded and spends more effort to detect the pupil without compromising real-time applicability. 
The main strategy which improves the algorithm against occlusions is extracting the elliptical arcs from input image and finding one arc or a group of arcs representing the pupil contour. 
In this way, relevant features from a partially visible pupil can be extracted and detection can be performed by fusion of separate features.
Besides detecting the pupil boundary and center precisely, the algorithm can also identify if there is no pupil, i.e. in case of a blink, in the image.
This information can be utilized in applications where users need to make selections on a gaze-based interaction interface or trigger events like clicking a button in an assistive eye tracking system for disabled just by blinking~\cite{Duchowski2002, De2009, Topal2014}. 

APPD follows a simple workflow and consists of the steps shown in Fig.~\ref{fig:SystemSchema}.
The processing pipeline starts by detection of the region of interest (ROI) by convolving the eye image with a Haar-like feature. 
Then, we extract edge segments each of which is a contiguous array of pixels.
The next step is to determine whether a \textit{near-circular} segment exists that traces the entire boundary of the pupil.
Such an edge segment would exist only if the pupil is clearly visible with no or very little occlusion.
To determine whether an edge segment has a circular geometry, we devise a fast heuristic method which utilizes gradient distribution of an edge segment.
On the condition that a near-circular segment is found, we extract elliptical arcs from only that segment.
If no near-circular segment is found, which would be the case if the pupil is severely occluded by eyelids or eyelashes, then arcs from all edge segments in the ROI are extracted.

Following the extraction of the elliptical arcs, we join them in every possible combination to generate a set of ellipse candidates that at least one of them traces the pupil boundary.
Candidates are finally evaluated for their relevance to be the actual pupil contour and the best one, if it exists, is chosen among the candidate ellipses.
In the following subsections we elaborate steps of APPD algorithm in more detail to make the discussion clear.

\subsection{ROI Estimation}
\label{sec:ROIEstimation}

\begin{figure}[!t]
	\centering
	\subfigure[]
	{\includegraphics[width=.99\linewidth]{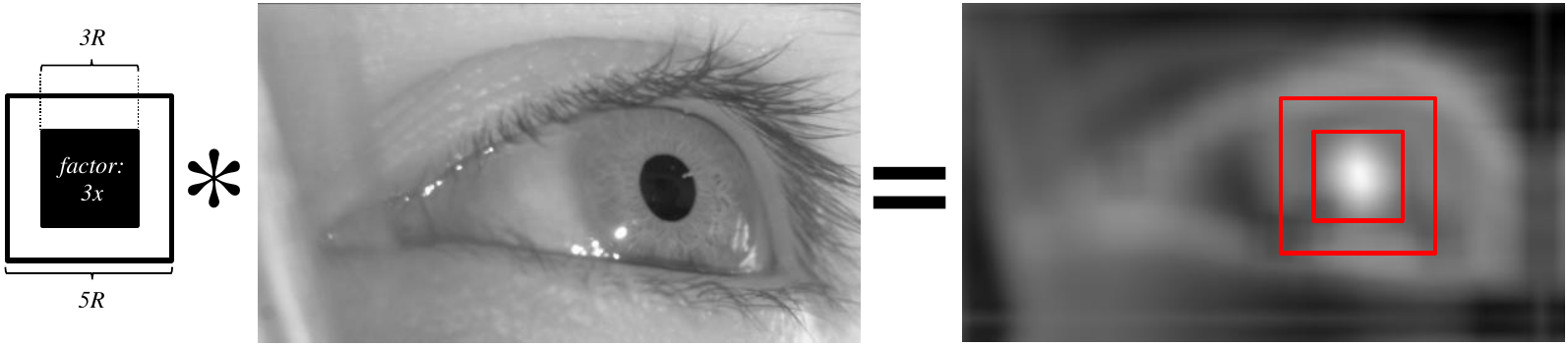}  
	\label{fig:ROIHaarFeature}}

	\subfigure[]{
		\includegraphics[width=0.495\linewidth]{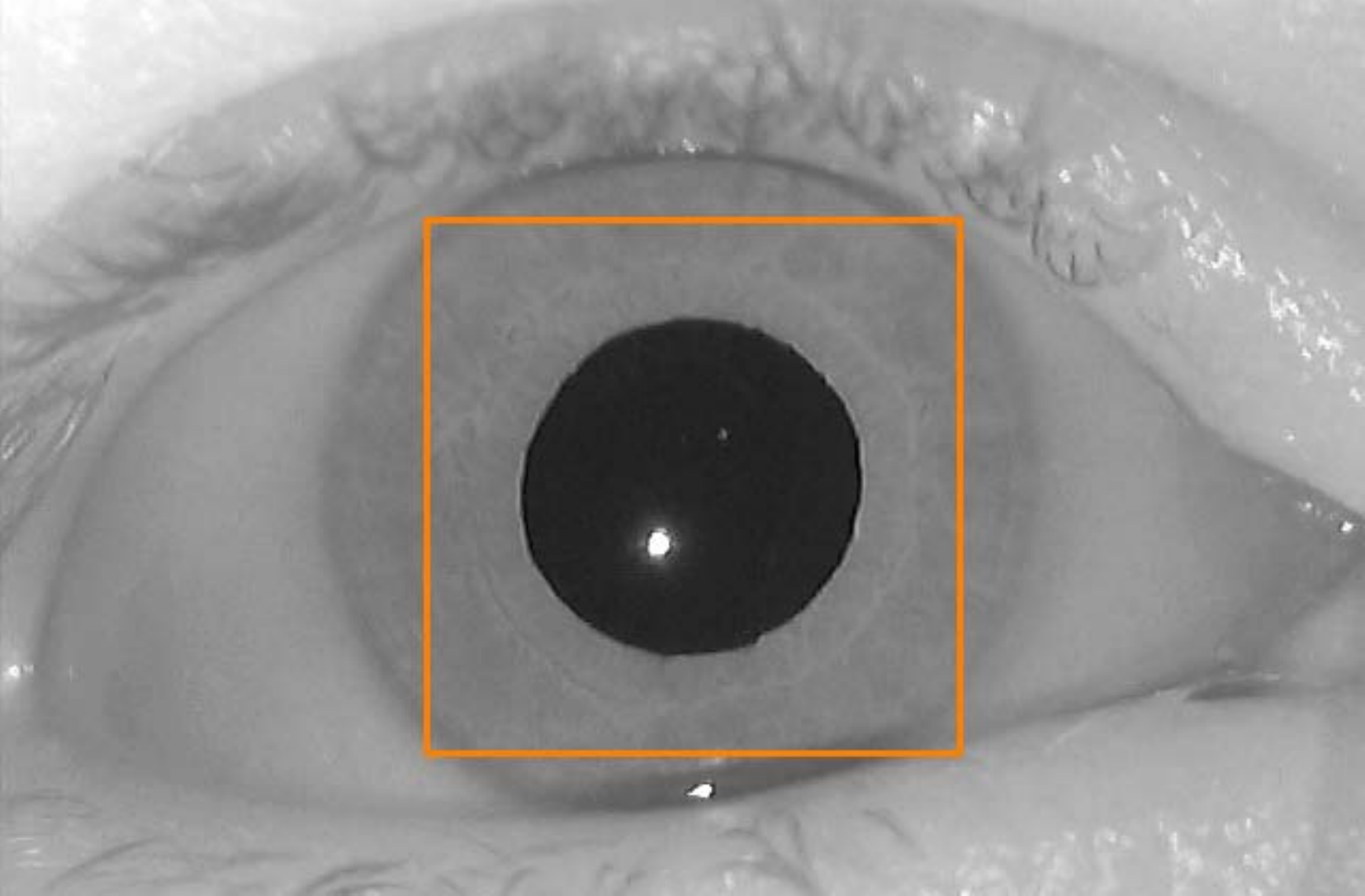} 
		\includegraphics[width=0.495\linewidth]{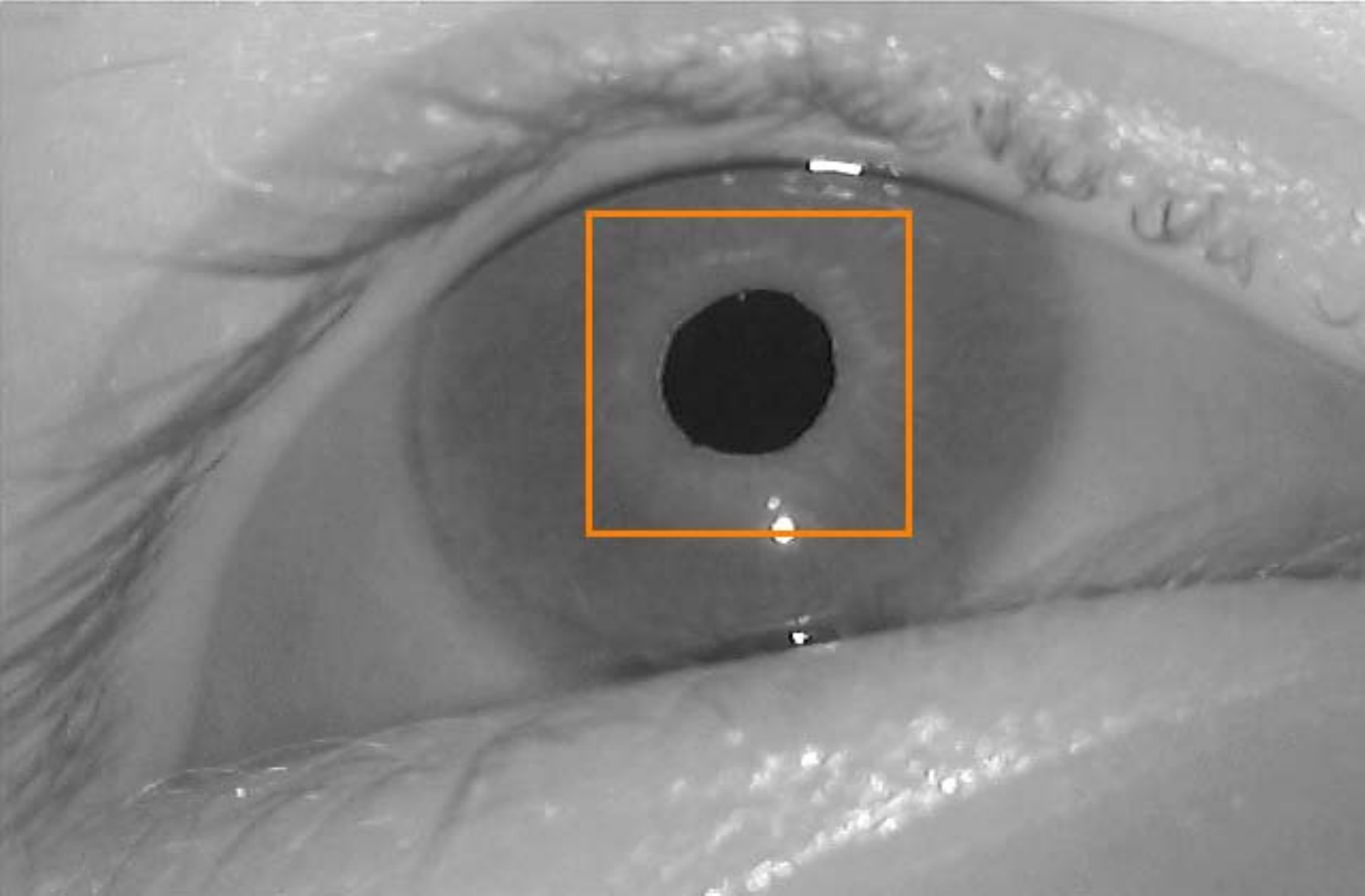} 
		\label{fig:LargeSmallROI}
	}
	\caption{(a) ROI detection by means of a convolution operation with a square Haar-like feature (left) and the convolution result (right). (b) Detected ROIs of two eye images having different size of pupils.}
	\label{fig:ROIEstimation}
\end{figure}

In the first step of APPD we roughly estimate the pupil area in the entire eye image.
For this purpose, we utilize pupil's geometric and intensity attributes in a similar vein as in~\cite{Swirski2012}.
A pupil can be described as a dark and compact blob since it consists of darker intensity levels than its surrounding iris, and it has usually an elliptical shape having a low eccentricity.
To locate pupil region we employ a square Haar-like feature~\cite{Viola2001} having a 3/5 ratio between inner and outer regions as seen in Fig.~\ref{fig:ROIHaarFeature}.

Human eye has several physiological properties that vary among the people such as eyeball radius, cornea curvature, distance between pupil center and cornea center, etc.~\cite{Hansen2010}.
Pupil size also varies among the population because of both the physiological differences among people, and also the dilation of pupil which occurs in light changes.

Therefore, we apply the Haar-like feature kernel in a multi-scale manner and pick the location where gives the maximum response per unit (i.e. total response divided by the total number of pixels in the kernel) at the end.
For this reason, we apply the Haar kernel in various aperture sizes from 150 px to 350 px with a step of 50 px, i.e. in five scales to find the pupil ROI in the images of resolution 1280$\times$720 px in our dataset.
In Fig.~\ref{fig:LargeSmallROI} results of ROI estimation process for two eye images from two individuals having different pupil sizes are presented.

\begin{figure}[!t]
	\centering{\includegraphics[width=\columnwidth]{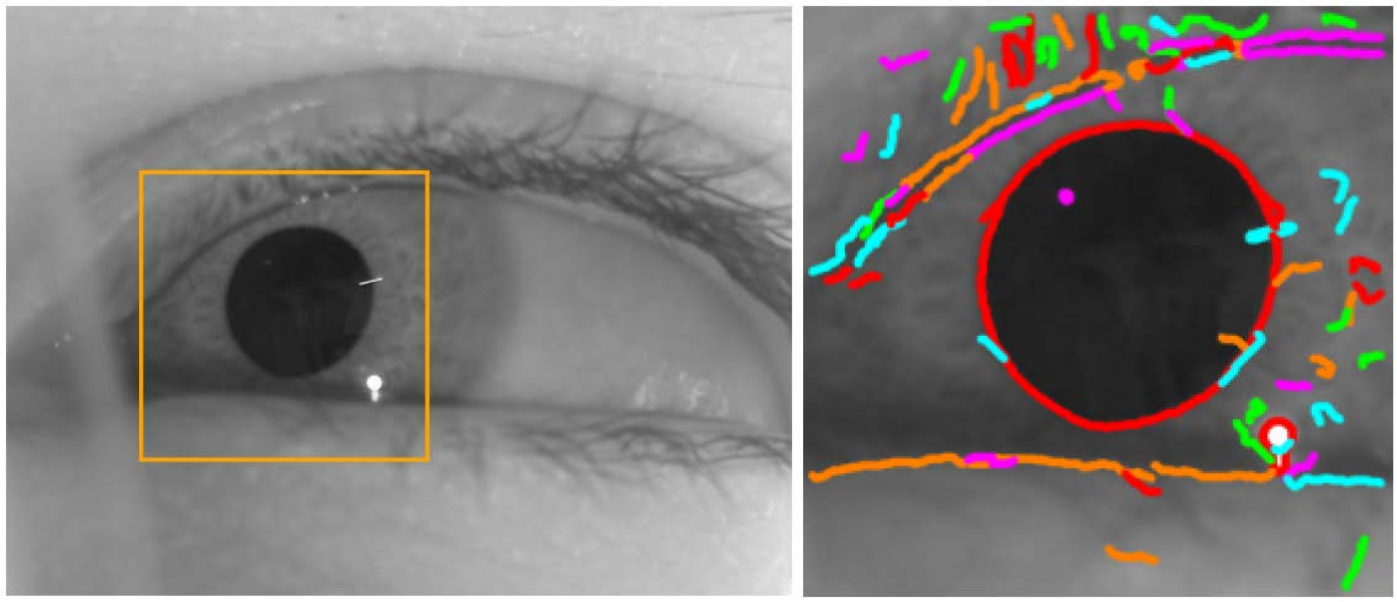}}  
	\caption{Sample eye image in which detected ROI is indicated (left), each edge segment obtained by EDPF algorithm within the ROI is indicated with an individual color (right).}
	\label{fig:EyeImageAndEdges}
\end{figure}

\subsection{Edge Segment Detection}
\label{sec:EdgeSegmentDetection}

To detect all edge segments inside the ROI, we employ Edge Drawing (ED) edge segment detector\footnote{\url{http://c-viz.anadolu.edu.tr/EdgeDrawing}}~\cite{Akinlar2013,Topal2012}. 
Unlike traditional edge detectors which work by identifying a set of potential edge pixels in an image and eliminating non-edge pixels through operations such as morphological operations, non-maximal suppression, hysteresis thresholding~\cite{Canny1986, Heath1997}, ED follows a proactive approach.
The ED algorithm works by first identifying a set of points in the image, called the anchors, and then joins these anchors in a way which maximizes the gradient response of edge paths, hence ensures good edge localization.
ED outputs not only a binary edge map similar to those output by conventional edge detectors, but it also outputs the result as a set of edge segments each of which is a contiguous and connected pixel chain~\cite{Topal2011}.
This property of ED extremely eases the application of the algorithm to further detection and recognition problems.

Similar to other edge detectors, ED has several parameters that must be tuned by the user for different tasks. 
Ideally, one would want to have an edge detector which runs with a fixed set of parameters for any type of image. 
To achieve this goal, we have incorporated ED with the \textit{a contrario} edge validation mechanism due to the Helmholtz principle~\cite{Desolneux2001, Desolneux2007}, and obtained a fast parameter-free edge segment detector, i.e. Edge Drawing Parameter-Free (EDPF)\footnote{\url{http://c-viz.anadolu.edu.tr/EDPF}}~\cite{Akinlar2012}. 

EDPF works by running ED with all ED's parameters at their extremes, which detects all possible edge segments in a given image with many false positives.
It then validates the extracted edge segments by the Helmholtz principle, which eliminates false detections leaving only perceptually meaningful segments with respect to the \textit{a contrario} approach with very little overhead of computation, i.e. a mere 1 ms for an HD frame.
Fig.~\ref{fig:EyeImageAndEdges} illustrates detected edge segments for an example eye image. 
In the figure, each color represents a different edge segment, which is one-pixel width, contiguous array of pixels.

\subsection{Near-circular Segment Search}
\label{sec:NearcircularSegmentSearch}

The main goal of this step is detecting the pupil in an easy and computation efficient way when its circumference is entirely visible in the case of no occlusion. 
Once we have the edge segments detected, we need to find the one that traverses the pupil boundary. 
The most intuitive solution is to apply a brute force search as follows: fit an ellipse to each edge segment, compute the fitting error, and pick the edge segment that has the smallest fitting error. 
This method might work only when the pupil is clearly visible, i.e., when it is not occluded by the glints of IR LEDs (infrared light emitting diodes) or eyelashes, however, fitting an ellipse and calculating fitting error for each segment requires too much computation. 

\begin{figure}[!t]
	\centerline{\includegraphics[width=\columnwidth]{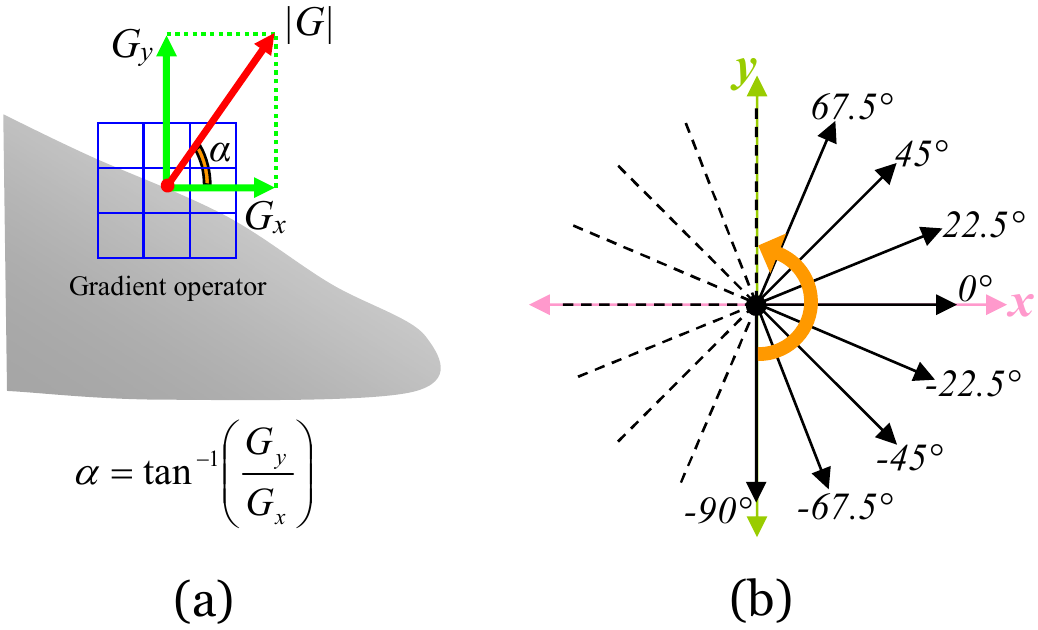}}  
	\caption{(a) Gradient computation with image derivatives, (b) Quantization of computed gradient directions.}
	\label{fig:GradientComputation}
\end{figure}

To reduce this computational burden, we devise a faster method based on the analysis of gradient directions to find the near-circular segment, if one exists.
Gradients of the segments contain substantial information about the geometrical structure and is used in shape matching, retrieval and recognition problems~\cite{Jia2000, Ortiz2010}. 
Since we already have the vertical and horizontal derivatives of eye image computed during the edge detection scheme, we can find the gradient directions by very less amount of computation (see Fig.~\ref{fig:GradientComputation}.a). 
%
The $arctan$ function obviously results angle values in an interval $[-\pi/2, \pi/2]$, providing an angle range of 180$^\circ$.
Before examining the distribution of gradients, we quantize the angles with 22.5$^\circ$ to obtain discrete symbols, thus we divide the unit circle into 16 regions into eight different directions (see Fig.~\ref{fig:GradientComputation}.b).

\begin{figure}[!t]
	\centering{\includegraphics[width=0.95\columnwidth]{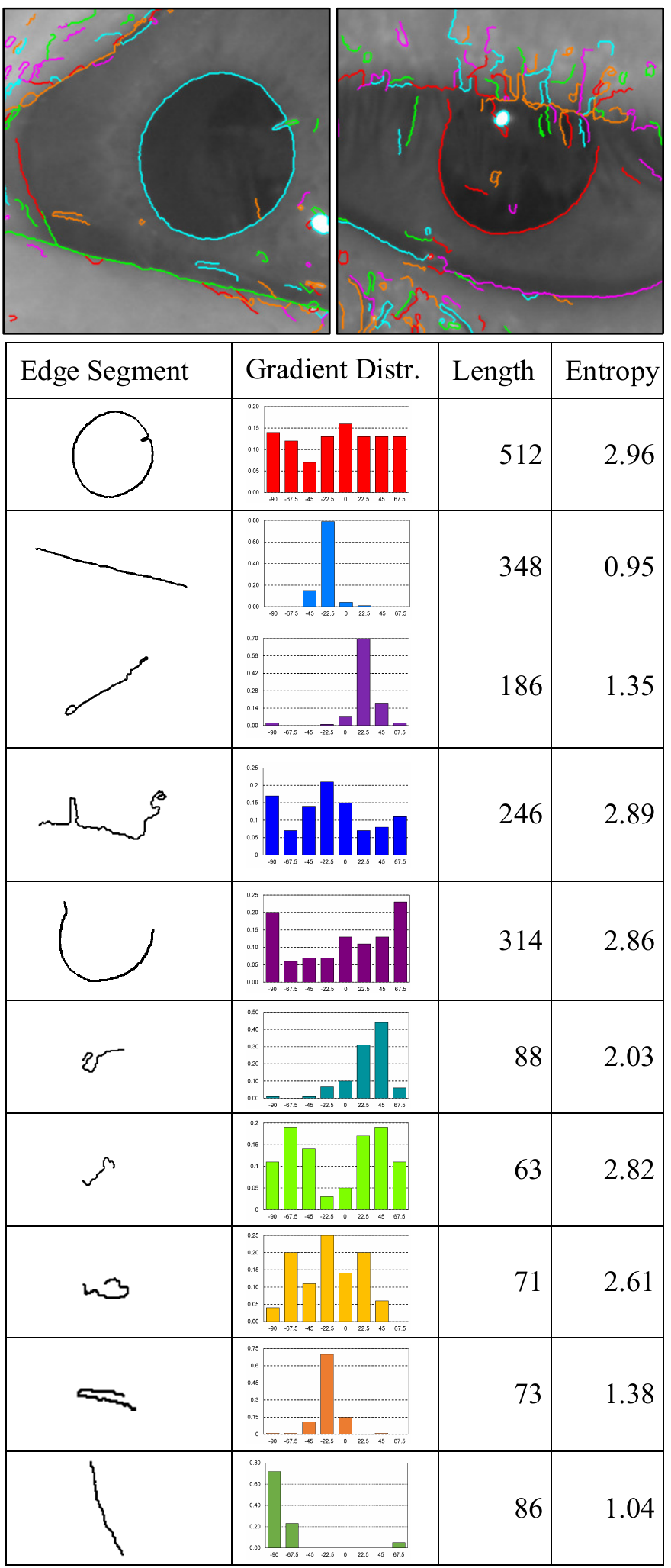}} 
	\caption{Two sample eye ROI with and without occlusion and detected edge segments (top),
		List of gradient distributions, lengths ans entropy value of several edge segments selected from above sample images (bottom).}
	\label{fig:SegmentsAndDistributions}
\end{figure}

Once we get quantized gradient directions for all pixels in a segment, we infer the shape characteristics of that segment by analysing their gradient distributions. 
It is easy to observe that any segment in near-circular form would have a uniform gradient distribution if the tangential gradients on its perimeter were sampled with a fixed angular step. 
Intuitively, circular edge segments would have relatively uniform gradient distribution; whereas, straight edge segments have an unbalanced distribution where a few values dominate. 
Thus, we can distinguish edge segments in circular shapes by picking the ones that result in a uniform gradient distribution. 
To achieve this, we use the entropy function (Eq.~\ref{eq:Eq_3_2_1}) on the quantized gradient distributions of the segments.
%
\begin{equation}
\label{eq:Eq_3_2_1} 
	E = -\sum_i^n p_i . \log(p_i)
\end{equation}
Since the entropy function maximizes for flat distributions where the frequency of each symbol is equal, we compute the entropy of gradient distribution for each separate edge segments as follows:
%
\begin{equation}
\label{eq:Eq_3_2_2} 
	\arg\max \left|\sum_i^8 f_{G_i} . \log(f_{G_i}) \right|
\end{equation}
where $f_{G_i}$ is the frequency of the $i^{th}$ gradient direction.
The entropy values for edge segments are maximized for a perfect circle and decreases as the segment shape differs from being a circle (elliptic, etc.), and finally entropy becomes zero for straight lines since a straight line has only one gradient direction along its trajectory.
Since we quantize the unit circle into eight directions (see Fig.~\ref{fig:GradientComputation}.b), the number of different symbols is eight and the maximum entropy value is $\log_2 8=3$ in our case.
Fig.~\ref{fig:SegmentsAndDistributions} shows edge segments of an input eye image and gradient distributions, lengths and entropy values for ten individual sample segments.
%
%
It is easy to observe that circular edge segments have higher gradient entropy values regardless their length, whereas straight edge segments have lower values as expected.
With this heuristic, we can discard the segments producing small entropy values than a certain threshold in an extremely fast manner. 

When examining speed of this method, we measure that computing the gradient entropy of an edge segment with available image derivatives is almost 200 times faster than ellipse fitting and error computation.
As a more explicit example, for an edge segment consisting of 500 pixels, ellipse fitting and error computation take 640 $\mu$s (190 $\mu$s and 450 $\mu$s, respectively) in total on an Intel 2.70 GHz CPU.
For the same segment, with available horizontal and vertical image derivatives, it only takes 3.5 $\mu$s to compute the entropy of segment gradients on the same machine.
In this way, we save significant amount of execution time by avoiding ellipse fitting and error computation for the segments that have irrelevant geometries rather than elliptical shapes. 

Once the gradient entropies of edge segments are computed, one segment is chosen to be the near-circular segment and elliptical arcs are extracted from it if the following three criteria are satisfied which we determine through empirical evaluation:
\begin{enumerate}[i)] 
\item Must have a high gradient entropy. The theoretical entropy upper-bound for eight different gradient directions is $\log_2 8=3$. Accordingly, we choose the segments which have 2.8 or more gradient entropy.
\item Must have a small ellipse fitting error, e.g., 2 pixels, when an ellipse is fit to the pixels which form the segment.
\item Must be a closed segment. To avoid problems due to the small occlusions such as glints, we consider a 15 pixels threshold for the distance between the start and end points of the segment.
\end{enumerate}

Along with the second criterion, ellipse fitting is an essential tool employed in various steps of the proposed method.
Among the many studies in the literature, there are two renowned ellipse fitting methods which are known to be fast and robust~\cite{Taubin1991},~\cite{Fitzgibbon1999}.
Among these two algorithms, Taubin's method~\cite{Taubin1991} results a better ellipse contour with slightly lower error, however, it does not guarantee that the resulted conic is an ellipse, rather it can return a hyperbola as well. 
In addition, Fitzgibbon's method~\cite{Fitzgibbon1999} always ensures that the resulted conic is an ellipse, but it tends to extract more eccentric ellipses with higher fitting errors.
To benefit advantages of both methods, we follow a simple Taubin-prior procedure as the following.
First, we use Taubin's method and examine the coefficients of resulted conic to understand whether its an ellipse or hyperbola. 
If it turns out that we get a hyperbola, then we use Fitzgibbon's method and get an ellipse.
Due to the fact that we apply ellipse fit to consecutive edge elements rather than scattered pixel data, we usually end up with a valid ellipse with Taubin's method.

In both ellipse fitting methods, we need to compute a fitting error to quantitatively evaluate the success.
For this purpose, there is no straightforward method in the literature except numerical approximations~\cite{Rosin1996}.
Since inaccurate approximations easily cause misjudgements of elliptical features, we developed a quantitative method~\cite{Cakir2018} based on~\cite{Nurnberg2006} to compute the fitting error precisely.
Once we estimate all distance values between each point and the ellipse, we calculate the root mean square error (RMSE) to obtain a single scalar to represent the fitting error.
Besides the fitting error computation, there is no straightforward method for ellipse perimeter computation which requires calculation of an infinite series for an exact solution.
Therefore, we employ Ramanujan's second approximation~\cite{Ramanujan1962} that is able to provide a very close estimate by a very handy computation.
The approximation for the ellipse perimeter $(P_e)$ with given semi-major ($a$) and semi-minor ($b$) axes is
%
\begin{equation}
\label{eq:Ramanujan_1} 
	P_e \approx \pi (a+b) \left( 1 + \frac{3h}{10 + \sqrt{4 - 3h}}\right) 
\end{equation}
where
\begin{equation}
\label{eq:Ramanujan_2} 
	h = \frac{(a-b)^2}{(a+b)^2}	
\end{equation}
%

In the event that more than one edge segment satisfies all three conditions given above, the one having the minimum ellipse fitting error is chosen to be the near-circular segment.
While existence of a near-circular segment speeds up the computation, its not compulsory for the detection of pupil.

It is important to note that shape of a segment does not necessarily have to be near-circular to give high entropy values.
In addition to segments with near-circular geometry, gradient distributions of segments which have concave shapes or follow complex trajectories can also end up with high entropy values.
Therefore, we use entropy test as a prerequisite to accelerate the algorithm and make final decision about a segment by ellipse fitting.

\begin{figure*}[!t]
	\centerline{\includegraphics[width=\textwidth]{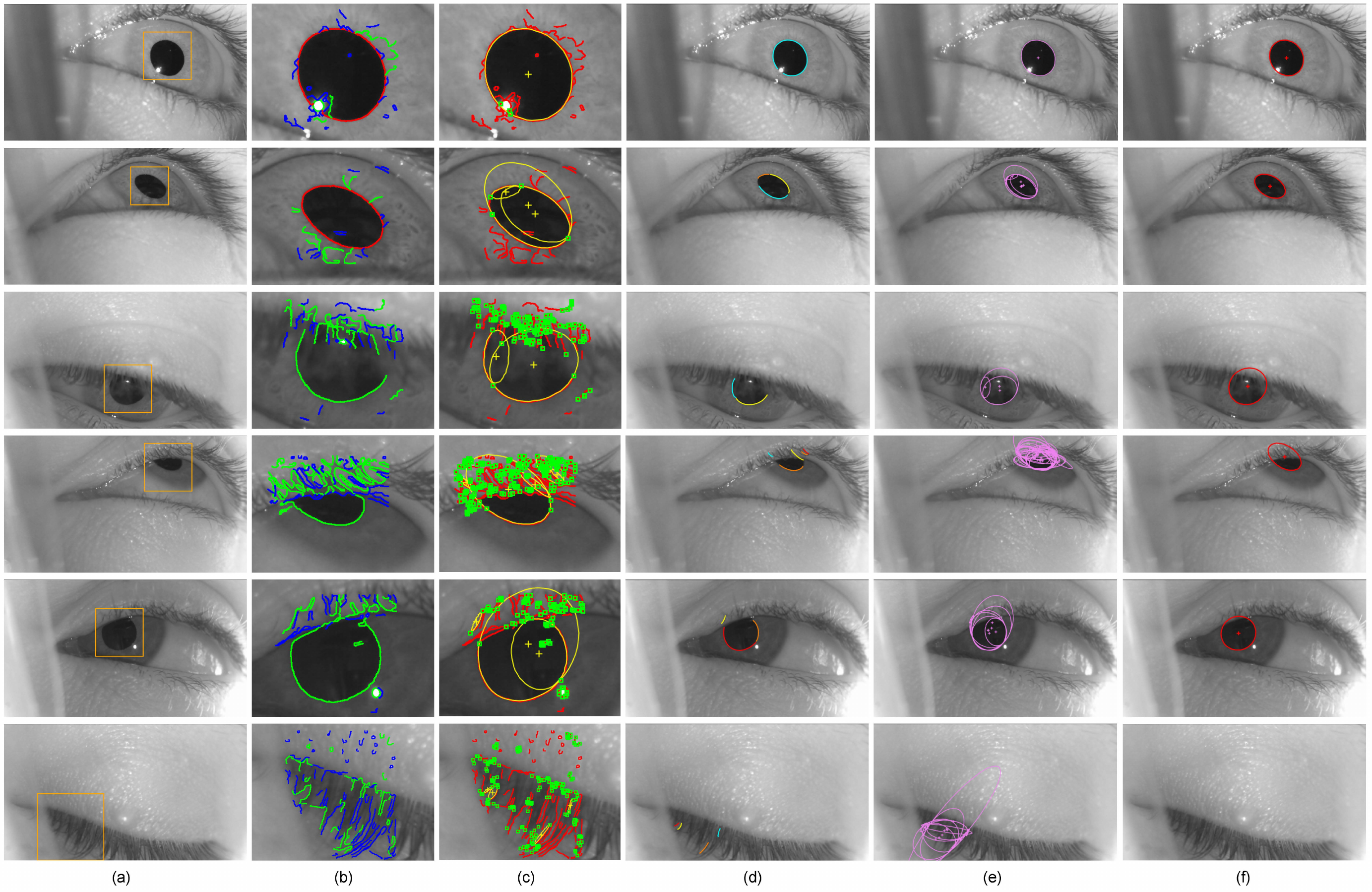}} 
	\caption{
		ROI detection, arc extraction, pupil candidate generation and pupil detection steps. 
		The first two rows consist of examples where the pupil is entirely visible; there are examples with occlusions in the rows 2-to-5, and there is no pupil in the last one. 	
		(a) Input image with detected ROI, 
		(b) Detected edge segments within the ROI. Near-circular segment is indicated in red, if exists. High entropy segments which are subjected to arc extraction if a near-circular segment could be found indicated in green. Short and low gradient entropy segments which are not omitted indicated in blue. 
		(c) Detected corners (green boxes) and ellipses which are fit to pixels lying in between two consecutive corners. Only successful (i.e. having low fitting error) ellipses are indicated.
		(d) Extracted elliptical arcs.
		(e) $2^n-1$ pupil candidates are generated by joining all possible arc combinations, 
		(f) the selected ellipse representing the pupil contour using the argument in Eq.~\ref{eq:CostFunction}	(Best viewed in color)
	}
	\label{fig:ArcExtractionAndPupilCandidateGeneration}
\end{figure*}

\subsection{Elliptical Arc Extraction}
\label{sec:EllipticalArcExtraction}

The next step of the algorithm is extracting the elliptical arcs (which will be referred to as \textit{arc} hereafter) from edge segments obtained in the previous step. 
If a near-circular segment could be found at the previous step, arcs are extracted only from that segment. 
If no near-circular segment is found, then all segments which have high gradient entropy (i.e., $>2$) are subjected to arc extraction process. 
In this manner, the algorithm adapts itself and requires less computation when there is no occlusions and pupil contour is entirely visible.
Due to the fact that their straight geometry rarely contains elliptical arcs, we omit segments having low gradient entropy and short segments (i.e., $<25$ pixels) to save further computation time.
 
In a previous work, we extract circular arcs by combining consecutive line segments to detect circles in an image\footnote{Demo page: \url{http://c-viz.anadolu.edu.tr/EDCircles}}~\cite{Akinlar2013}.
However, pupil's projection onto the camera plane can be more elliptic, hence we need to detect the elliptical arcs in this study.
To solve this problem, we devise another strategy that finds the start and end points of a potential elliptical arc within an edge segment by locating the corners along the segment~\cite{Cakir2016}.
We detect corners on the segments with a fast curvature scale-space (CSS) method which utilize image gradient information to compute turning angle curvature~\cite{Topal2013}.
Curvature is a function that indicates the amount by which a geometric entity (an edge contour in our case) deviate from being planar.
Along an edge contour, the curvature function gives higher responses where sharp changes on trajectory occurs.
In this manner, corner locations can be extracted efficiently by only processing edge segments instead of entire 2D image.

Afterwards, we apply ellipse fit to the points lying in between two consecutive corners along each segment and obtain elliptical arcs.

In Fig.~\ref{fig:ArcExtractionAndPupilCandidateGeneration}.b-d we present the results of arc extraction process for several test images that is shown in Fig.~\ref{fig:ArcExtractionAndPupilCandidateGeneration}.a with the extrcted ROI.
In the first two rows, the pupil is completely visible; hence, the near-circular segment (indicated in red) is detected.
Therefore, arc extraction is applied to only this segment. 
When no near-circular segment is found due to occlusions, arcs are extracted from all segments having high gradient entropy to avoid missing any critical information (see $3^{rd}$, $4^{th}$ and $5^{th}$ rows of Fig.~\ref{fig:ArcExtractionAndPupilCandidateGeneration}).

We should note that the pupil it may appear highly elliptical due to an oblique view angle and thus the gradient distribution of the edge segment enclosing the pupil may not provide a uniform distribution even though there is no occlusion.
In such cases, the gradient distribution results a low gradient entropy, therefore near-circular segment might not be detected. 
Consequently, elliptical arc extraction would be applied to all segments although the pupil is entirely visible.

\begin{figure}[!t]
	\centering{\includegraphics[width=\columnwidth]{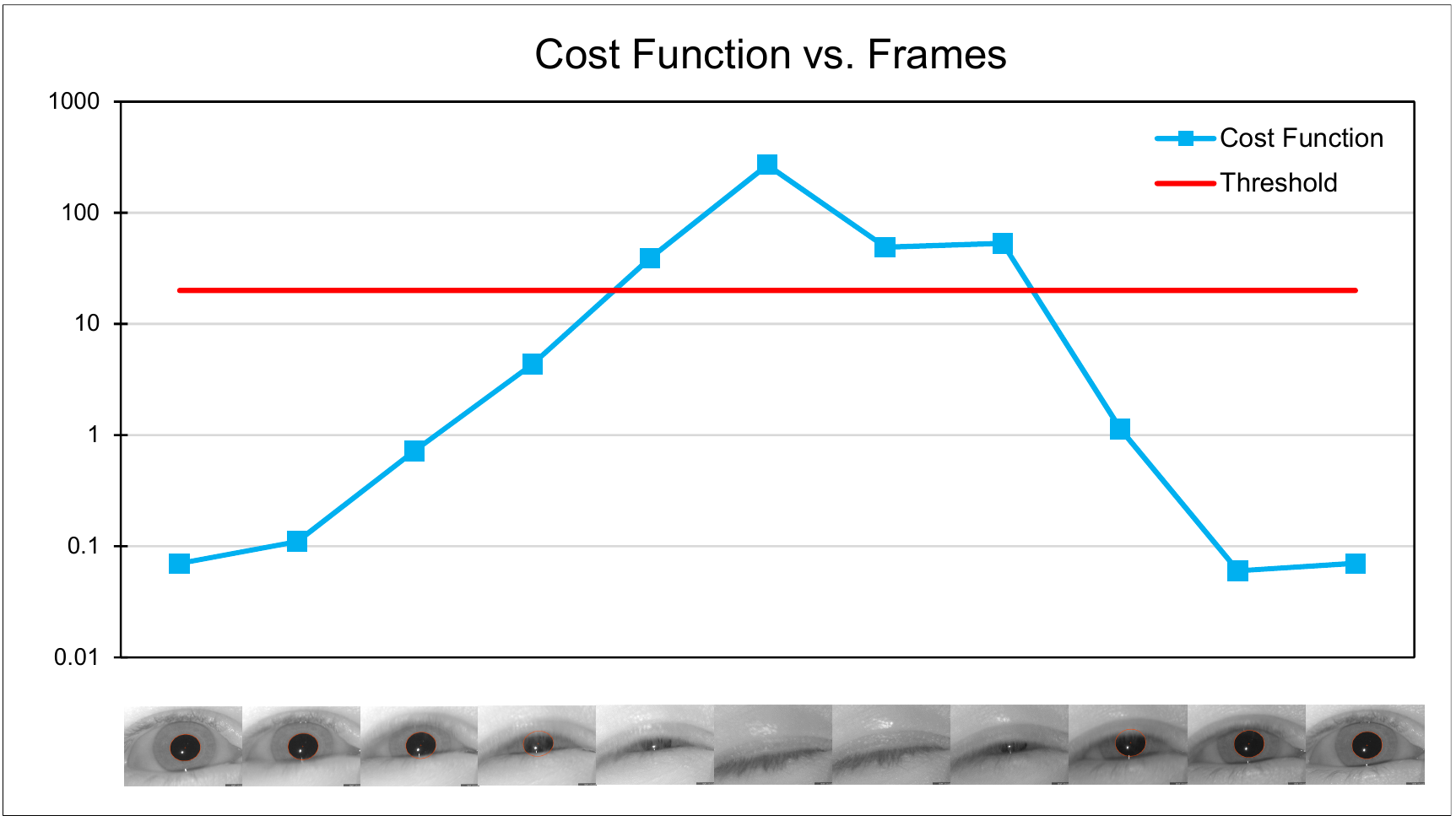}} 
	\caption{Output of our cost function for a set of example frames. It is seen that the result of cost function increases proportional to the pupil occlusion. If the occlusion is dramatic or there is no pupil in the image, the value of the function overshoots. (Note that the plot is in logarithmic scale.)}
	\label{fig:CostFunctionPlot}
\end{figure}

\subsection{Pupil Candidate Generation}
\label{sec:GeneratingPupilCandidates}

In this step, we generate candidate ellipses by grouping the extracted arcs.
Thus we aim to detect the pupil boundary completely even if its boundary is partially visible.
To generate pupil candidates, we try to fit ellipse to each subset of all extracted arcs. 
Excluding the empty set, there are $2^n-1$ different arc combinations for $n$ arcs.
Fig.~\ref{fig:ArcExtractionAndPupilCandidateGeneration}.e shows all generated pupil candidates generated from extracted arcs in Fig.~\ref{fig:ArcExtractionAndPupilCandidateGeneration}.d.

Since the pupil candidate generation process considers all subsets of selected arcs, groups of unrelated arcs which do not form a valid elliptic structure are also subjected to be eliminated after ellipse fit.
Therefore, we eliminate those candidates which result high fitting error due to the fact hat they cannot belong to the pupil boundary.
After we eliminate candidates which result high fitting error, one of the remaining candidates is going to be selected as the final pupil by utilization of a cost function in the final step.
Or, the algorithm ends up with the decision that there is no pupil in the image, if the output of the cost function diverges.

\begin{figure}[!t]
	\centering
	\includegraphics[width=0.475\linewidth]{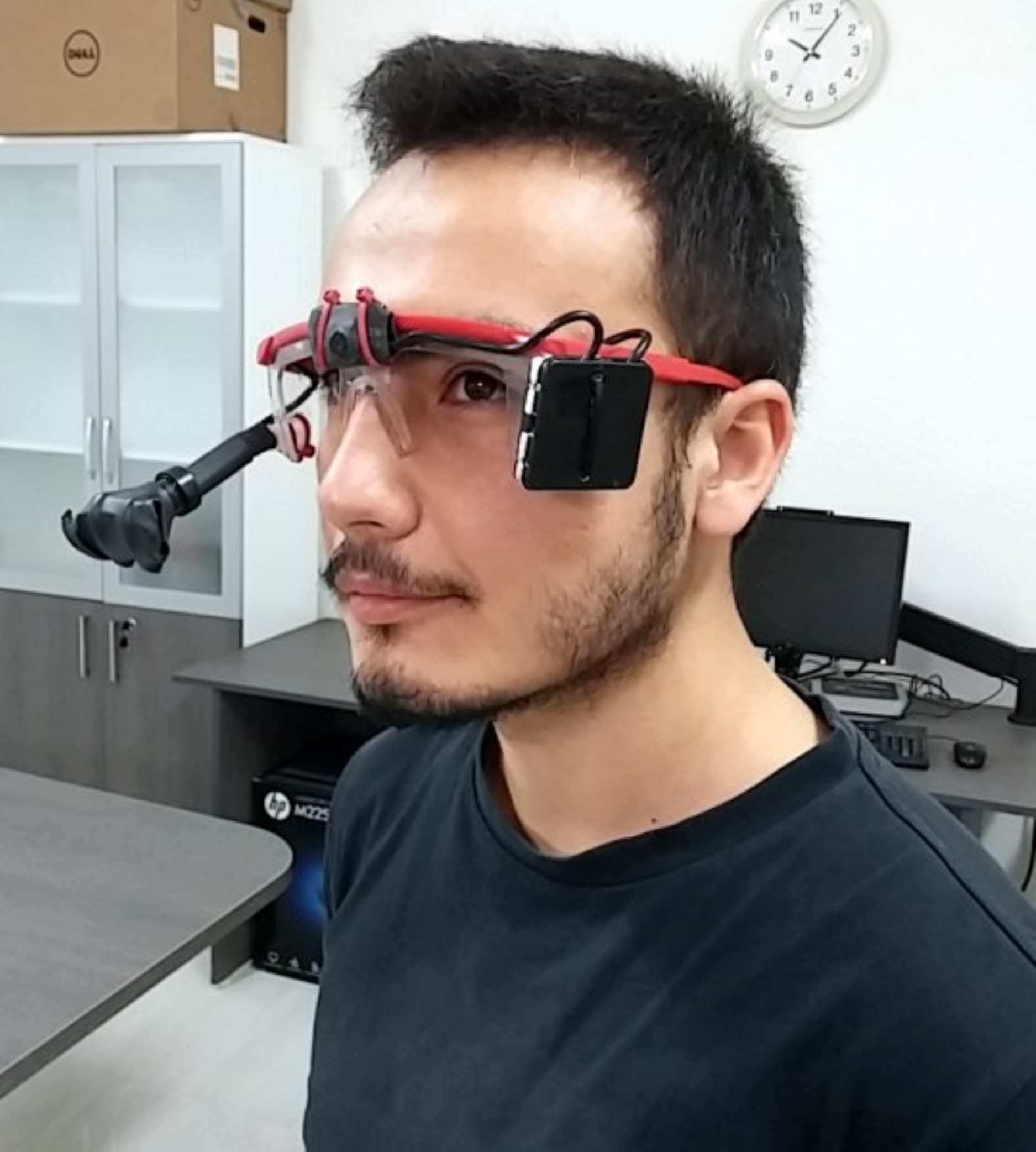}  
	\includegraphics[width=0.475\linewidth]{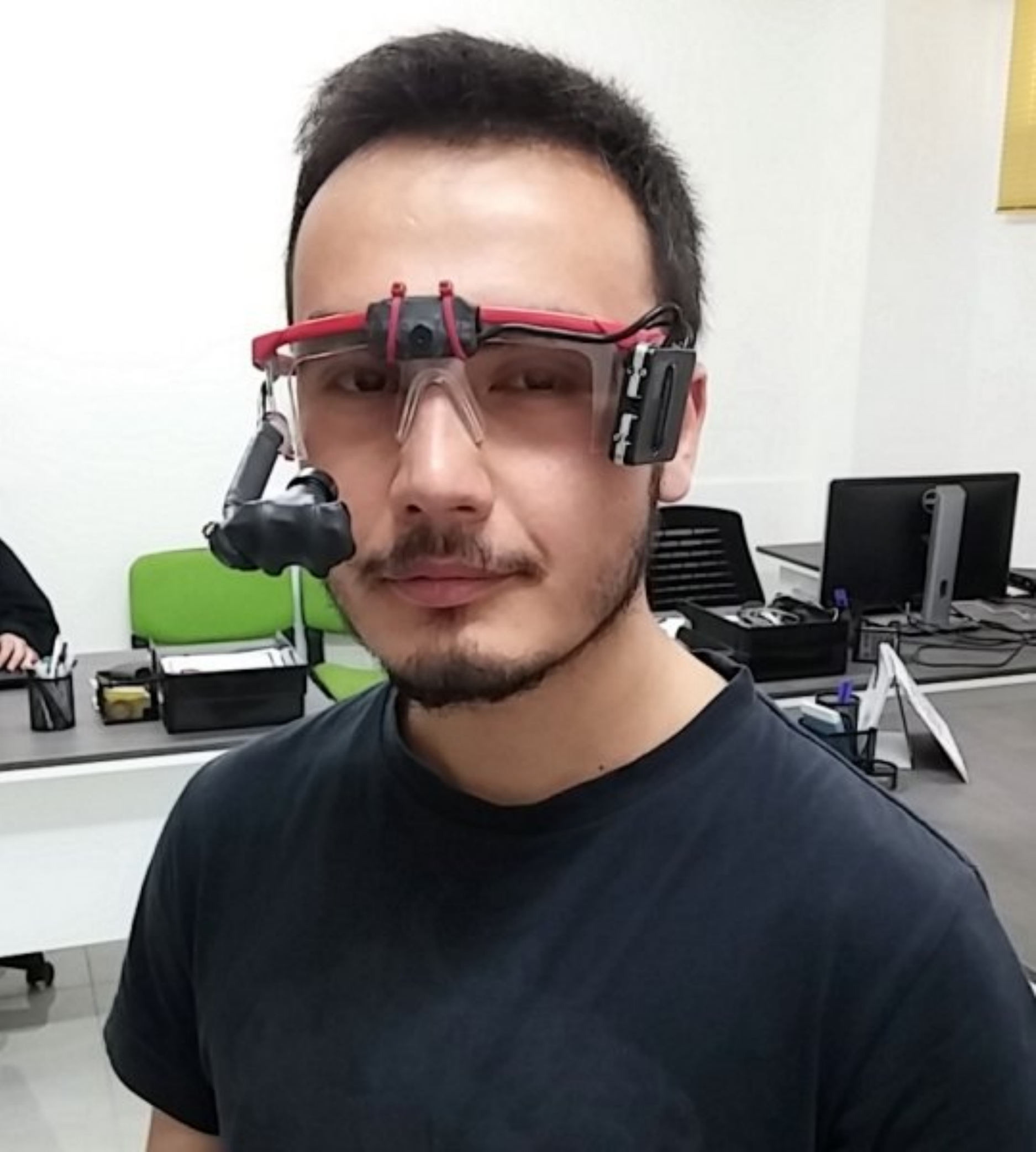}  
	\caption{Images of apparatus that we collect database video sequences.}
	\label{fig:AparatFoto}
\end{figure}

\subsection{Detection of the Pupil}
\label{sec:DetectionOfPupil}

In the previous step, we get a number of pupil candidates each of which is a subset of elliptical arcs consisting several arcs.
Accordingly, we still need to select one candidate ellipse to be the final pupil contour.
To make the decision, we define a cost function $J_c$ which considers the following properties of a candidate ellipse:
\begin{enumerate}[i.] \itemsep0em
\item the ellipse fitting error ($\varepsilon$), 
\item the eccentricity ($e$),
\item the ratio of the arc pixels to the perimeter of resulting ellipse ($\phi$).
\end{enumerate}
Each of the pupil candidates is formed by one or more arcs.
If the pupil boundary is detected from multiple arcs, the fitting error should be reasonable because we expect the arcs to be parts of the same elliptic contour.
Thus we need to minimize the fitting error $\varepsilon$. 

The eccentricity ($e$) indicates the compactness of an ellipse, or in other words, diversity of an ellipse from a circle and it is computed as
%
\begin{equation}
\label{eq:eccentricity}  
e = \sqrt{1 - \frac{b^2}{a^2}} \;, \;\;\;\;\;\; 0<e<1
\end{equation}
where $a$ and $b$ are semi-major and semi-minor axes, respectively.
The eccentricity is zero for a circle and one for a parabola. 
Among the pupil candidates each of which is a subset of elliptical arcs, there are also diverse ellipses whose eccentricities can get close to 1. 
However, pupil's projection onto the image plane is usually closer to a circle rather than a skewed ellipse in majority of the applications.
Therefore, we tend to select a candidate having an eccentricity close to 0. 

The parameter $\phi$ is the ratio of the number of pixels involved in ellipse fitting to the perimeter of the resulting ellipse.
In some circumstances, one single and short arc may result in a large pupil candidate ellipse that may lead to inconsistency. 
Therefore, we look for the pupil candidates which are formed by consensus of more arc pixels and have a greater $\phi$.

During the experiments, we observed that the effect of the eccentricity ($e$) is less than the effect of $\varepsilon$ and $\phi$ due to the possibility of true pupil not being the most compact ellipse among the candidates.
Accordingly, we take squares of $\varepsilon$ and $\phi$ to increase their effect on the cost function.
Finally, we need to minimize $\varepsilon$ and $e$ and maximize $\phi$ in our formulation, and select the candidate that minimizes the following argument:
%
\begin{equation}
\label{eq:CostFunction} 
    J_c(p_i)= {\arg\min}_{(\varepsilon,e,\phi)} \left| \frac{\varepsilon_i^2 \,.\, \pi^e_i}{\phi_i^2} \right|		
\end{equation}
where $p_i$ is the $i^{th}$ pupil candidate and $\pi$ is constant.

Fig.~\ref{fig:ArcExtractionAndPupilCandidateGeneration}.f shows the pupil detection results for sample images.
Among the pupil candidates shown in Fig.~\ref{fig:ArcExtractionAndPupilCandidateGeneration}.e, the one that minimizes the $J_c$ in Eq.~\ref{eq:CostFunction} is selected as the pupil.

\begin{figure}[!t]
	\centering{\includegraphics[width=\columnwidth]{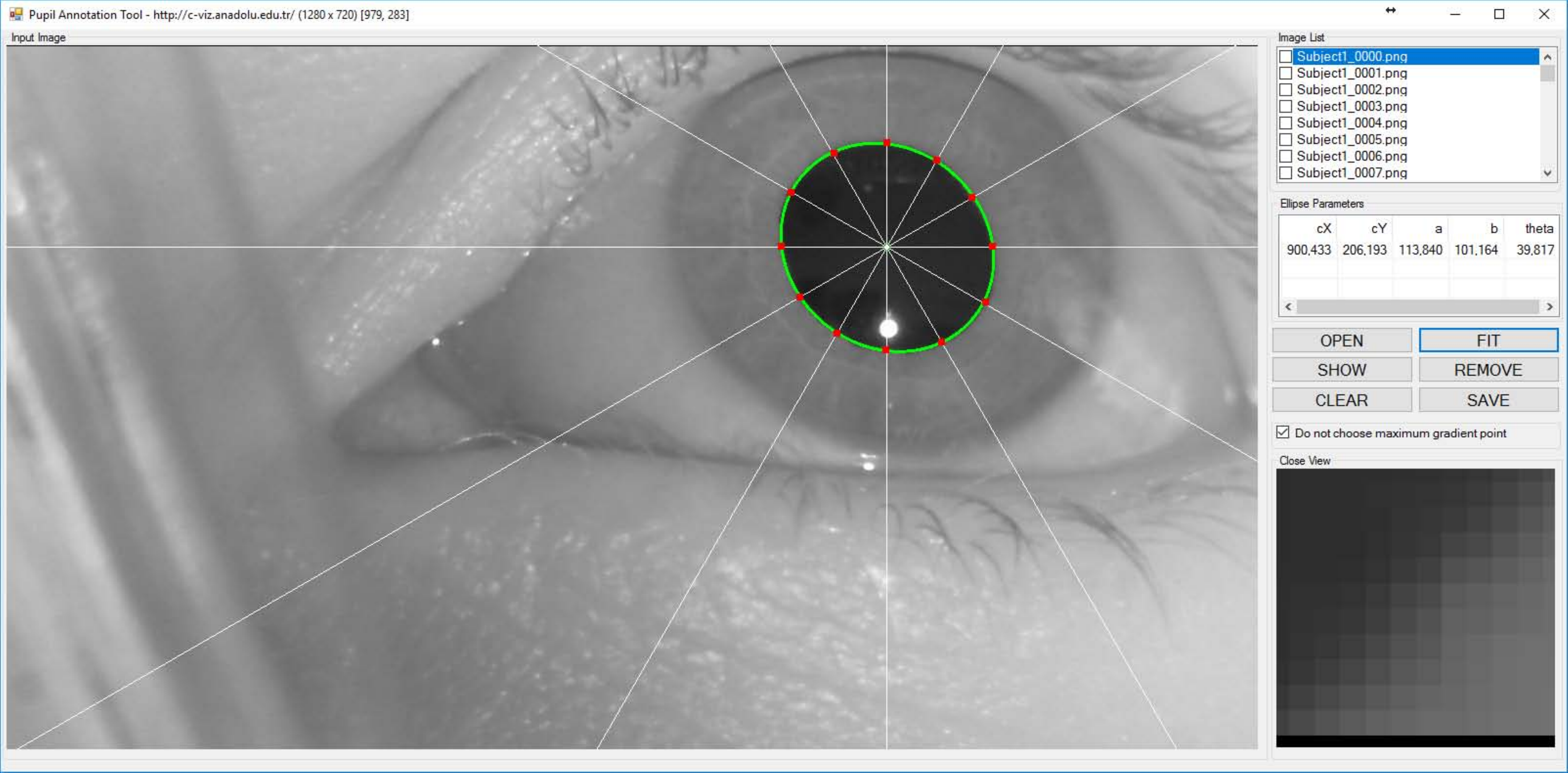}} 
	\caption{A snapshot of our pupil annotation tool. 
		Two different ellipse fitting algorithms are utilized to find the best conic to represent the pupil GT. 
 		Once clicking a location inside the pupil, a guide (white lines) is displayed in order to help user to equally sample contour points.
 		When the annotator starts to select pupil boundary pixels, the application makes a local search around the clicked location and picks the coordinates with the greatest gradient magnitude to make sure the correct boundary pixels are selected.}
	\label{fig:PupilAnnotationTool}
\end{figure}

\subsection{Detection of True Negatives}
\label{sec:DetectionOfTrueNegatives}

In many applications, having the information that there is no pupil in the image is important as much as detecting it.
This information can provide very useful extensions to eye tracking applications such as blink detection.
In our algorithm, it is still possible to obtain arcs and pupil candidates although there is actually no pupil. 
We observe that the cost function overshoots in these circumstances due to large $\varepsilon$ and small $\phi$ values. Therefore, we can easily find if there is no pupil by quantifying output of $J_c$. 
In Fig.~\ref{fig:CostFunctionPlot} we present a plot of the cost function versus a number of frames sampled from an eye blink operation.

It is clearly seen that the $J_c$'s output rapidly increases as the visible part of the pupil periphery gets smaller due to occlusions.
Similarly, the algorithm ends up that there is no pupil in the image because the cost function overshoots for all of pupil candidates in the last row of Fig.~\ref{fig:ArcExtractionAndPupilCandidateGeneration}.
By examining several frames, we find out that a stable threshold value can provide promising results on deciding whether there is no pupil.
We provide more detail on this topic in the next section with quantitative experimental results.

\begin{figure*}[!t]
	\centerline{\includegraphics[width=1\textwidth]{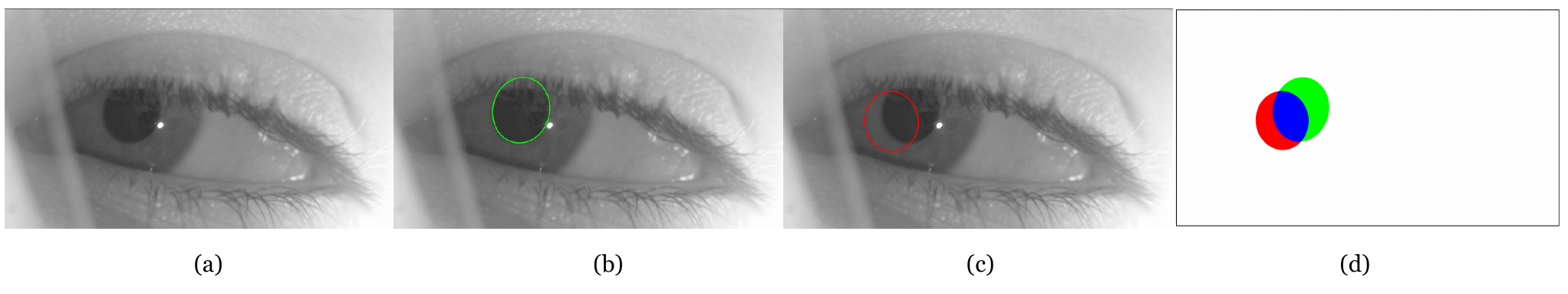}} 
	\caption{Illustration of localization test. (a) Input image, (b) Pupil ground truth pupil (GT), (c) Detected pupil (Det), (d) Overlapping (indicated in blue) and non-overlapping (indicated in red and green) pixels.}
	\label{fig:LocalizationPixels}
\end{figure*}

\section{Experimental Results}
\label{sec:ExperimentalResults}

In this section, we present results of a comprehensive set of experiments in both quantitative and qualitative manner. 
We compare APPD to three state of the art pupil detection algorithms, i.e. Starburst~\cite{Li2005}, \'{S}wirski et al.~\cite{Swirski2012} and ElSe~\cite{Fuhl2016}.
We quantitatively assess algorithms in terms of pupil detection accuracy (by means of F-measure), pupil localization and running time.
We also provide qualitative results that provide useful clues to readers about the performance of the algorithms.
In addition to the content that we present in the paper, we also provide more supplementary material (e.g. codes, videos, etc.) on our website~\cite{APPDWebsite}.

\subsection{Pupil Detection Dataset}
\label{sec:Dataset}

In order to perform experiments, we first prepare a dataset containing 3904 high resolution (1280$\times$720 px) eye frames collected from 12 subjects.
We used a simple head-mounted eye tracking apparatus (see Fig.~\ref{fig:AparatFoto}) consisting of two HD cameras (for scene \& eye) that we built for 3D gaze estimation study. 
To our knowledge, this is the only available pupil detection dataset in resolution higher than VGA (640$\times$480 px) in the literature.

During collection of the frames, we ask all subjects to move their eyes different directions in a certain order. 
In this way, we obtain eye images as pupil is viewed in diverse angles with and without occlusions by camera.
Furthermore, we also want users to blink several times to obtain negative image samples that pupil does not exist.
Eventually, in 57\% of the frames pupil is entirely visible, in 22\% of them there are severe occlusions and in 21\% of them there is no pupil in the dataset.
We count a pupil positive sample if more than half of its periphery is visible, otherwise it is considered as a negative sample. 

After we collect the test frames, we implement an efficient annotation tool which eases rigorous annotation procedure (see Fig.~\ref{fig:PupilAnnotationTool}). 
Our annotation tool overlays a grid in polar coordinates to ease the selection of pixels from pupil boundary in an equal angular resolution.
In addition, to ensure localization of ground truth (GT) conics, it does not collect the exact pixel coordinates that users click.
Instead, it searches a local pixel neighborhood of clicked location to find the maximum image gradient response and picks that location.
In this way, we guarantee the selection of exact edge pixels in between pupil and iris in high resolution images precisely.

Even though having five points is sufficient to fit an ellipse hence its degree of freedom, we picked ten points in average from each pupil's boundary to better reduce the effect of perspective distortion.
Note that a circle's projection onto image may not be a perfect ellipse due to the perspective distortion and lens distortion.
Once all points are set along the pupil's boundary, then we fit ellipse to them with two different algorithms~\cite{Taubin1991,Fitzgibbon1999} and select the parameters which provides lower fitting error. 
Therefore, we obtain the best possible conic to represent the pupil GT in each eye image.
%

\subsection{Localization Assessment}
\label{sec:LocalizationEvaluation}
The first quantitative test we perform is localization assessment of pupil detection algorithms.
In this evaluation, we quantify success of algorithms for how precisely they detect pupils with respect to ground truth data.
We apply this test only the frames in which the pupil is truly detected by each algorithm in the dataset.
Source codes of all algorithms were downloaded from the websites that authors provided in the corresponding papers
\footnote{Starburst source codes:\\ \url{https://github.com/thirtysixthspan/Starburst}}
\footnote{\'{S}wirski's source codes:\\ \url{https://github.com/LeszekSwirski/pupiltracker/}}
\footnote{ElSe source codes:\\ \url{ftp://messor.informatik.uni-tuebingen.de/}}. 
We set all parameters of all algorithms according to their corresponding publications, or use the best performing values if it is not explicitly indicated in the paper or code. 
For each algorithm, we used a single parameter set for all images in the dataset which provides the best overall result.
We present the parameter listings for all algorithms in Table~\ref{Parameters}. 

\begin{table}[!b]
	\centering
	\caption{Parameter listing of the algorithms employed in the experiments.}
	\resizebox{1\columnwidth}{!} 
	{
		\begin{tabular}{l|l}
			\toprule
			\textbf{Algorithm} & \textbf{Parameters} \\ \midrule
			\parbox{15mm}{Starburst\\MATLAB} & 
			\parbox{70mm}{Window Size = 301\\Max. RANSAC Iterations = 10000\\Number of Rays = 200} 
			\\ \hline
			
			\parbox{15mm}{\'Swirski\\C/C++} &   
			\parbox{70mm}{\vspace{1.5mm} Canny Low Thr.= 30, Canny High Thr.= 50\\Haar: Min. Radius = 45, Max. Radius = 130\\Number of RANSAC Iterations = 30\\Early Term Percentage = 95\%} 
			\\ \hline
			
			\parbox{15mm}{\vspace{1mm} ElSe\\C/C++} &  
			\parbox{70mm}{\vspace{1.5mm} Validity Thr. = 10, Neighborhood Size = 2\\Min. Area = 0.5\%, Max. Area = 10\%} 
			\\ \hline
			
			\parbox{15mm}{APPD\\C/C++} &  
			\parbox{70mm}{\vspace{1.5mm} Gradient Entropy Threshold = 2.9\\Haar: Min. Aperture = 150 px\\Max. Aperture = 350 px, Step = 50} 
			\\ \bottomrule
		\end{tabular}
	}
	\label{Parameters}	
\end{table}

In order to quantify the localization performance, we compute overlap ratio (\textsc{OR}) between the detected pupil and ground truth by counting number of corresponding pixels as follows:
%
\begin{equation}	
	\textsc{OR}(E_{Det},E_{GT}) = \frac{Area(E_{Det}) \cap Area(E_{GT})}{Area(E_{Det}) \cup Area(E_{GT})}
	\label{eq:OverlapRatio} 
\end{equation}
where $E_{Det}$ and $E_{GT}$ are ellipse of detected pupil and ground truth ellipse, respectively~\cite{Prasad2012}.
Note that the range of \textsc{OR} is $[0,1]$ where it is zero and one for non-overlapping and perfectly overlapping ellipses, respectively.
In this manner we calculate the ratio of the number of overlapping pixels to total number of overlapping and non-overlapping pixels as seen in Fig.~\ref{fig:LocalizationPixels}. 

%
%
\begin{figure*}[!t]
	\centerline{\includegraphics[width=1.05\textwidth]{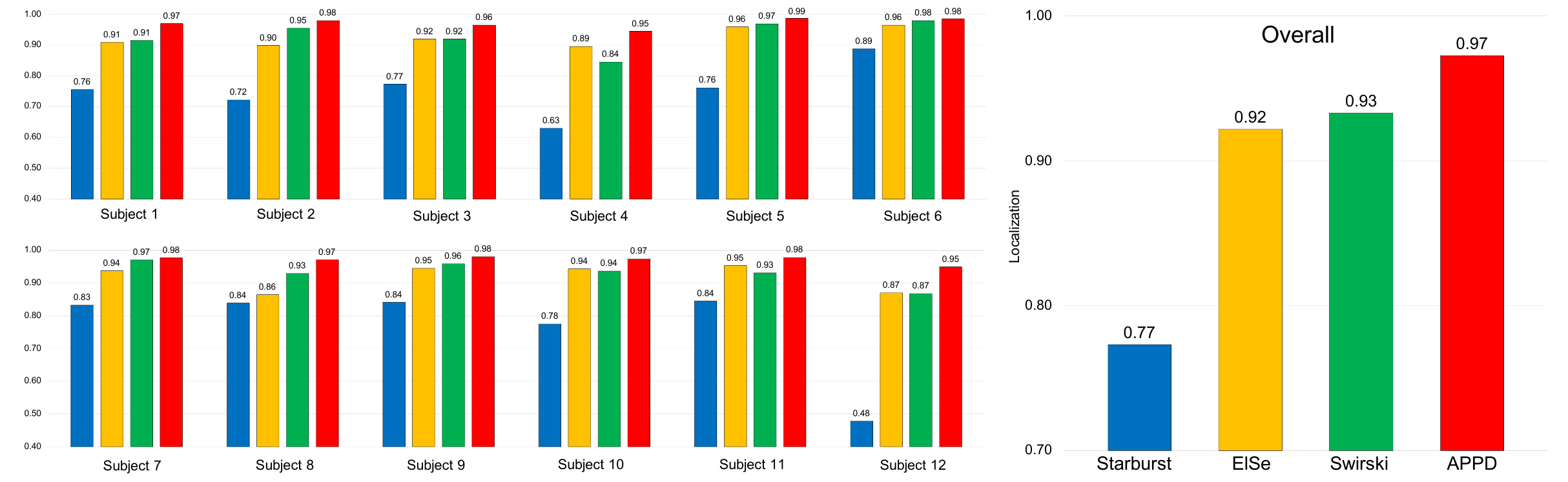}} 
	\caption{Localization test results of each algorithm for each subject.}
	\label{fig:LocalizationChart}
\end{figure*}

Once the number of overlapping and non-overlapping pixels are determined, we calculate \textsc{OR} and take the average for all images for each algorithm.
Higher \textsc{OR} indicates better localization, hence provides higher accuracy in the application which pupil detection is utilized, obviously.
In Fig.~\ref{fig:LocalizationChart} we present average localization results for individual subjects and overall for each algorithm.
Although ElSe and \'{S}wirski also give promising results, APPD algorithm performs the best (97\%) with a 4\% improvement over the runner up (93\%) in overall results.
\subsection{Accuracy Assessment}
\label{sec:AccuracyEvaluation}
In the previous experiment we assess the localization performance of algorithms by considering only the images that they detect pupil correctly. 
In this step, we assess the accuracy of algorithms by counting the number of images that the pupil is correctly detected in the entire frame sequences.
To consider a pupil image as a correct detection (TP), we calculate overlap error ($\varepsilon_O$) as in Eq.~\ref{eq:OverlapError} and compare the result with a threshold value~\cite{Chia2011}. 
\begin{equation}
	\label{eq:OverlapError} 
	\varepsilon_O(E_{Det},E_{GT}) = 1 - \textsc{OR}
\end{equation}
The range of $\varepsilon_O$ is in between 0 and 1, and its value obviously increases as the intersection area between detected ellipse and GT decreases.
We compare $\varepsilon_O$ with a threshold value to make a decision on the detected pupil on whether it is a true positive (TP) or a false positive (FP).
Likewise, we also evaluate images in which algorithms do not detect a pupil as true negative (TN) if there is no actual pupil in the image; or false negative (FN) vice versa.
After we count TP, FP, TN and FN samples, we calculate Precision (Eq.~\ref{eq:Precision}) and Recall (Eq.~\ref{eq:Recall}) values in order to compute F-Measure (Eq.~\ref{eq:Fmeasure}).

\begin{figure*}[!t]
	\centerline{\includegraphics[width=1.05\textwidth]{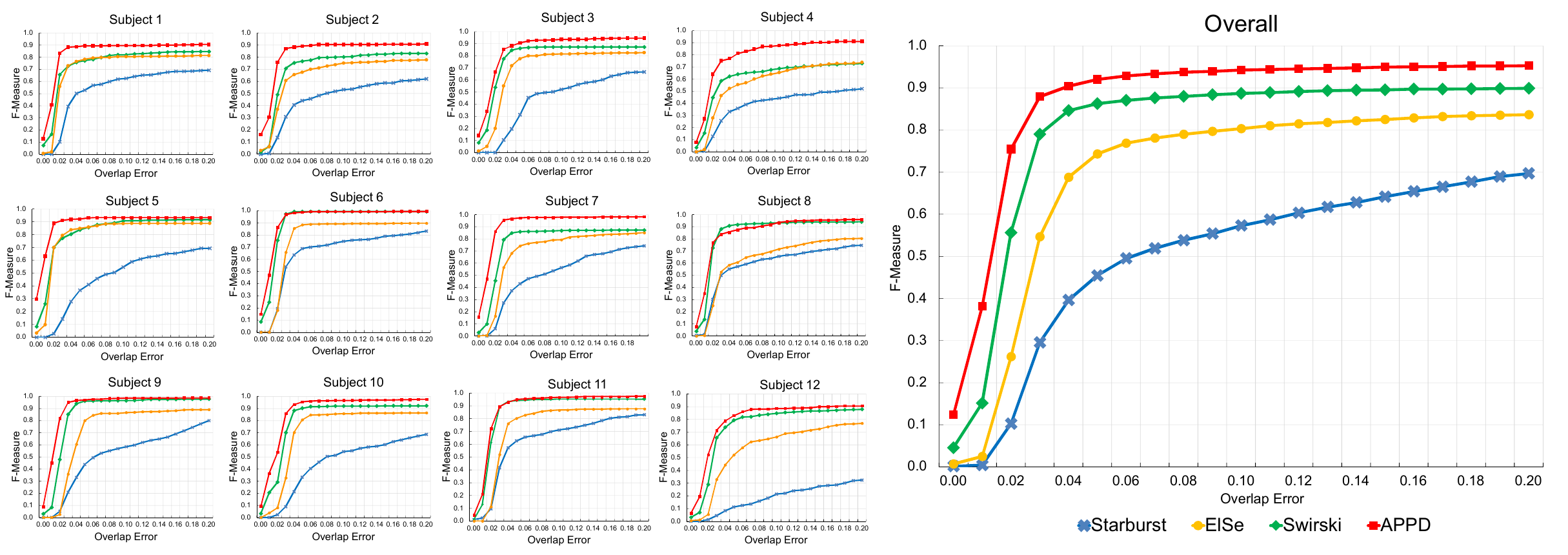}} 
	\caption{F-Measure results of each algorithm for each subject.}
	\label{fig:AccuracyChart}
\end{figure*}

We present F-Measure results in Fig.~\ref{fig:AccuracyChart} with respect to a range of $\varepsilon_O$ in between 0.0 to 0.2 which corresponds to \textsc{OR} varying from a perfectly aligned ellipses at 80\% overlap.
We do not consider pupils as TP if they are detected with a \textsc{OR} lower than 80\%.
From the sketches in Fig.~\ref{fig:AccuracyChart} it is clearly seen that accuracy tests are less contentious than localization experiments where the performances of the algorithms are closer. 
In this experiment, APPD outperforms others as its accuracy rapidly increases even in very small $\varepsilon_O$ errors and follows a very stable path regardless the subject.
We also see that \'{S}wirski and ElSe algorithms performs very closely with a notable success over Starburst algorithm.
%
%
\begin{equation}
\label{eq:Precision}  
\textit{Precision} = \frac{\textit{count{(TP Pupils)}}}{count{(\textit{TP Pupils} + \textit{FP Pupils})}}
\end{equation}
%
\begin{equation}
\label{eq:Recall}
\textit{Recall} = \frac{\textit{count{(TP Pupils)}}}{count{(\textit{TP Pupils} + \textit{FN Pupils})}}
\end{equation}
%
\begin{equation}
\label{eq:Fmeasure}
\textit{F-Measure} = \frac{2 \times \textit{Precision} \times \textit{Recall}}{\textit{Precision} + \textit{Recall}}
\end{equation}
%
%

\subsection{Qualitative Results}
\label{sec:QualitativeResults}

\begin{figure*}[!t]
	\centerline{\includegraphics[width=.97\textwidth]{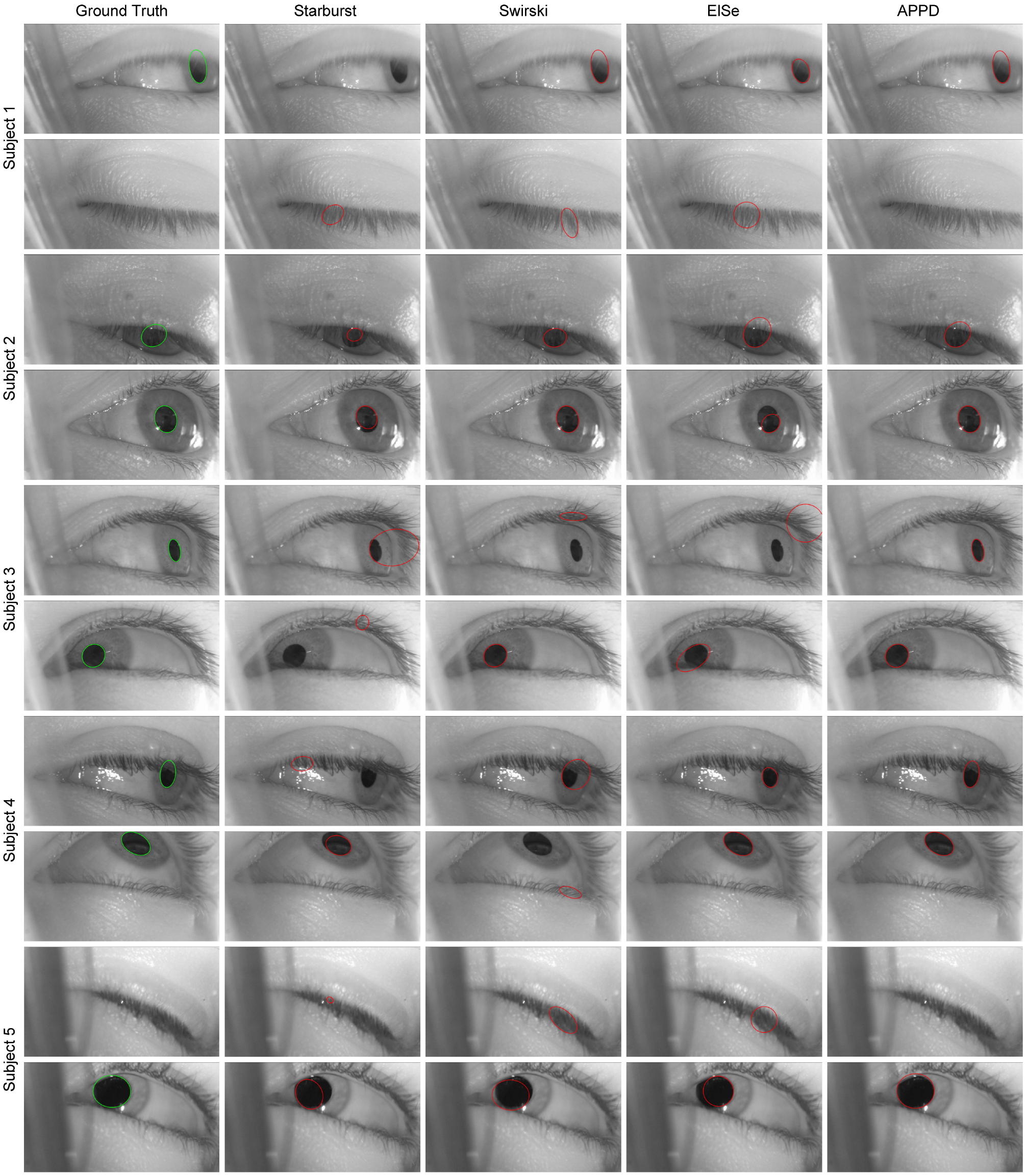}} 
	\caption{Qualitative results from all algorithms and all subjects (images in every two-rows from the beginning belong to a different subject).}
	\label{fig:QualitativeResults}
\end{figure*}

\begin{figure*}[!t]
	\centerline{\includegraphics[width=1\textwidth]{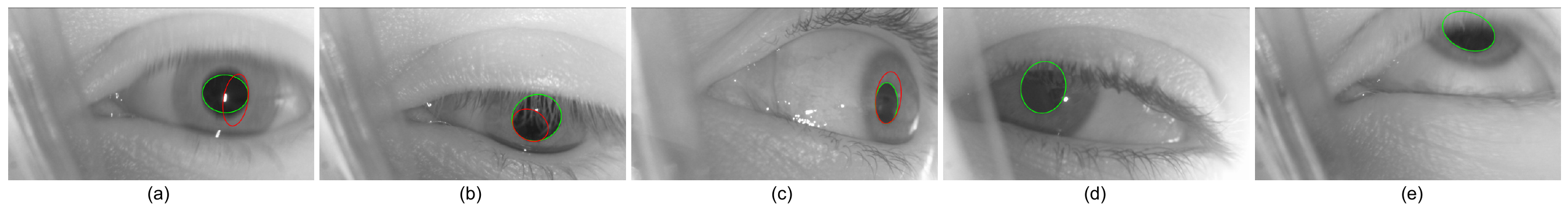}} 
	\caption{Several examples in which our algorithm fails. The algorithm could detect no pupil in the last two images.}
	\label{fig:QualitativeFails}
\end{figure*}

Along with quantitative accuracy and localization results, we also present qualitative results in Fig.~\ref{fig:QualitativeResults}. 
In the figure, we provide two results from each of 12 subjects from top to bottom.
It is also clearly shown that the algorithm can successfully determine true negatives, i.e. in 2$^{nd}$ and 9$^{th}$ rows.
In Fig.~\ref{fig:QualitativeFails} we present several examples where APPD fails.
The most common reason for fail cases is motion blur where the algorithm cannot extract edges from the pupil contour.
Therefore, elliptical arcs, hence the pupil contour cannot be detected.
Besides the images presented here, we also provide video sequences of all algorithms in~\cite{APPDWebsite} for interested readers.

%
%

\begin{table}[!b]
	\centering
	\caption{Average running times of algorithms for each subject in milliseconds. Best timings are indicated in \textbf{bold}.}
	\scalebox{0.85} 
	{
		\begin{tabular}{lrrrrrr}
			\toprule
			& \multicolumn{4}{c}{\textbf{Algorithm}}\\ \cmidrule{2-5}
			\textbf{Subject} & \multicolumn{1}{c}{\textbf{Starburst}} & \multicolumn{1}{c}{\textbf{\'{S}wirski}} & \multicolumn{1}{c}{\textbf{ElSe}} & \multicolumn{1}{c}{\textbf{APPD}} \\ \midrule
			{Subject 1}      & 88.84  & 50.94  & 7.22 & \textbf{6.02} \\
			{Subject 2}      & 29.96  & 47.10  & 7.28 & \textbf{5.32} \\
			{Subject 3}      & 41.42  & 47.11  & 7.13 & \textbf{4.30} \\
			{Subject 4}      & 48.87  & 41.47  & \textbf{7.51} & 9.14 \\
			{Subject 5}      & 195.63 & 55.70  & 6.78 & \textbf{5.42} \\ 
			{Subject 6}      & 26.75  & 55.89  & 7.04 & \textbf{4.43} \\
			{Subject 7}      & 37.39  & 51.82  & 7.06 & \textbf{5.11} \\
			{Subject 8}      & 59.36  & 50.22  & 7.18 & \textbf{6.92} \\
			{Subject 9}      & 36.27  & 44.79  & 6.98 & \textbf{4.11} \\
			{Subject 10}     & 10.89  & 38.12  & 6.85 & \textbf{3.45} \\ 
			{Subject 11}     & 24.21  & 49.96  & 6.92 & \textbf{5.31} \\
			{Subject 12}     & 14.09  & 35.27  & 7.48 & \textbf{4.82} \\ \hline
			\textbf{Average} & 49.22  & 47.17  & 7.12 & \textbf{5.37} \\ \bottomrule
		\end{tabular}
	}
	\label{TimingResults}	
\end{table}

\begin{table}[!b]
	\centering
	\caption{Dissection of average timing results of APPD algorithm for different subjects.}
	\scalebox{0.8} 
	{
		\begin{tabular}{lrrrrrrr}
			\toprule
			& \multicolumn{6}{c}{\textbf{Algorithm Step}} \\ \cmidrule{2-7}			 
			\textbf{Dataset} & \begin{sideways} \parbox[t]{20mm}{{ROI}\\{Detection}} \end{sideways} & \begin{sideways} \parbox[t]{24mm}{{Edge Segment} \\ {Detection}} \end{sideways} & \begin{sideways} \parbox[t]{24mm}{{Gradient}\\{Entropy Comp.}} \end{sideways} & \begin{sideways} \parbox[t]{24mm}{{Corner}\\{Detection}} \end{sideways} & \begin{sideways} \parbox[t]{24mm}{{Arc} \\ {Extraction}} \end{sideways} & \begin{sideways} \parbox[t]{24mm}{{Pupil}\\{Detection}} \end{sideways} & \begin{sideways}\parbox[t]{20mm}{\textbf{TOTAL (ms)}} \end{sideways} \\ \midrule
			Subject 1 &	\multicolumn{1}{c}{2.78} & \multicolumn{1}{c}{1.67} & \multicolumn{1}{c}{0.29} & \multicolumn{1}{c}{0.06} & \multicolumn{1}{c}{0.52} & \multicolumn{1}{c}{0.70} & \multicolumn{1}{r}{\textbf{6.02}} 
			\\
			Subject 2 &	\multicolumn{1}{c}{2.47} & \multicolumn{1}{c}{1.30} & \multicolumn{1}{c}{0.21} & \multicolumn{1}{c}{0.07} & \multicolumn{1}{c}{0.43} & \multicolumn{1}{c}{0.84} & \multicolumn{1}{r}{\textbf{5.32}} 
			\\
			Subject 3 &	\multicolumn{1}{c}{2.28} & \multicolumn{1}{c}{1.12} & \multicolumn{1}{c}{0.19} & \multicolumn{1}{c}{0.04} & \multicolumn{1}{c}{0.28} & \multicolumn{1}{c}{0.39} & \multicolumn{1}{r}{\textbf{4.30}} 
			\\
			Subject 4 &	\multicolumn{1}{c}{2.71} & \multicolumn{1}{c}{1.71} & \multicolumn{1}{c}{0.17} & \multicolumn{1}{c}{0.11} & \multicolumn{1}{c}{0.58} & \multicolumn{1}{c}{3.86} & \multicolumn{1}{r}{\textbf{9.14}}  
			\\
			Subject 5 &	\multicolumn{1}{c}{2.60} & \multicolumn{1}{c}{1.60} & \multicolumn{1}{c}{0.26} & \multicolumn{1}{c}{0.06} & \multicolumn{1}{c}{0.39} & \multicolumn{1}{c}{0.51} & \multicolumn{1}{r}{\textbf{5.42}} 
			\\
			Subject 6 &	\multicolumn{1}{c}{2.26} & \multicolumn{1}{c}{1.03} & \multicolumn{1}{c}{0.29} & \multicolumn{1}{c}{0.04} & \multicolumn{1}{c}{0.30} & \multicolumn{1}{c}{0.51} & \multicolumn{1}{r}{\textbf{4.43}} 
			\\
			Subject 7 &	\multicolumn{1}{c}{2.50} & \multicolumn{1}{c}{1.48} & \multicolumn{1}{c}{0.18} & \multicolumn{1}{c}{0.09} & \multicolumn{1}{c}{0.36} & \multicolumn{1}{c}{0.50} & \multicolumn{1}{r}{\textbf{5.11}} 
			\\
			Subject 8 &	\multicolumn{1}{c}{3.14} & \multicolumn{1}{c}{2.06} & \multicolumn{1}{c}{0.25} & \multicolumn{1}{c}{0.07} & \multicolumn{1}{c}{0.44} & \multicolumn{1}{c}{0.96} & \multicolumn{1}{r}{\textbf{6.92}} 
			\\
			Subject 9 &	\multicolumn{1}{c}{2.16} & \multicolumn{1}{c}{0.98} & \multicolumn{1}{c}{0.18} & \multicolumn{1}{c}{0.03} & \multicolumn{1}{c}{0.27} & \multicolumn{1}{c}{0.48} & \multicolumn{1}{r}{\textbf{4.11}} 
			\\
			Subject 10 &	\multicolumn{1}{c}{1.86} & \multicolumn{1}{c}{0.69} & \multicolumn{1}{c}{0.17} & \multicolumn{1}{c}{0.03} & \multicolumn{1}{c}{0.23} & \multicolumn{1}{c}{0.47} & \multicolumn{1}{r}{\textbf{3.45}} 
			\\
			Subject 11 &	\multicolumn{1}{c}{2.54} & \multicolumn{1}{c}{1.23} & \multicolumn{1}{c}{0.18} & \multicolumn{1}{c}{0.05} & \multicolumn{1}{c}{0.36} & \multicolumn{1}{c}{0.95} & \multicolumn{1}{r}{\textbf{5.31}} 
			\\
			Subject 12 &	\multicolumn{1}{c}{2.50} & \multicolumn{1}{c}{1.40} & \multicolumn{1}{c}{0.10} & \multicolumn{1}{c}{0.05} & \multicolumn{1}{c}{0.24} & \multicolumn{1}{c}{0.53} & \multicolumn{1}{r}{\textbf{4.82}} 
			\\ \hline
			\textbf{Overall} & \multicolumn{1}{c}{2.48} & \multicolumn{1}{c}{1.36} & \multicolumn{1}{c}{0.21} & \multicolumn{1}{c}{0.06} & \multicolumn{1}{c}{0.37} & \multicolumn{1}{c}{0.89} & \multicolumn{1}{r}{\textbf{5.37}} 
			\\ \bottomrule	
		\end{tabular}
	}
	\label{DetailedTiming}	
\end{table}

\subsection{Running Time Assessment}
\label{sec:RunningTimeAssessment}

We run all experiments on a laptop computer with Intel i7 2.70 GHz CPU.
To be able to make a fair comparison, we take implementation platforms into consideration.
All algorithm implementations are in C++ except Starburst which is in MATLAB.
According to~\cite{Fornaciari2014}, a typical execution in MATLAB is 50 times slower than a C++ based application, therefore we divide timing results of Starburst by 50.
\'{S}wirski's implementation was implemented in order to benefit from parallel computing libraries in order to utilize multi-core CPUs.
To able to make a fair comparison, we assign the application to a specific core and measure running times.
The running times of all algorithms in average for all images are summarized in Table~\ref{TimingResults}.

According to average running times in Table~\ref{TimingResults}, APPD is the fastest one among all algorithms. 
It is shown that the APPD algorithm can run up to ~185 Hz in single thread for HD images in 1280$\times$720 px resolution on a 2.70 GHz CPU.
In per subject analysis we see that APPD is slightly slower than ElSe algorithm for Subject 4.
When we investigate the reason behind the longer execution of Subject 4, we see that there are too many occlusions which cause algorithm to fail on detecting a near-circular segment and extract arcs from all edge segments.

Table~\ref{DetailedTiming} gives a dissection of running times of APPD algorithm for individual steps. 
ROI detection is obviously the most computation demanding step of the algorithm which takes roughly half of the entire execution due to the computation of integral images and convolution of Haar-like features at several scales.
Another time-consuming step -especially when a near-circular segment could not be detected- is the last step, i.e. pupil detection, which is the main reason behind the algorithm's fall back at subject 4.
Since this step contains too many computationally expensive ellipse fitting and error calculation routines, it significantly stretchs out the execution time if absence of a near-circular segment is the case.

\subsubsection{Effect of Adaptiveness}
\label{sec:EffectofAdaptiveness}

We also investigated the computational gain provided by the adaptiveness property of the algorithm.
We separately compute the average running times for the images with and without occlusion in our dataset.
According to the experiments, we measure that APPD takes a mere $3.82$ ms and $7.35$ ms for images with and without occlusions respectively.
This experiment shows that the algorithm can save up to 48\% of computation time by detecting only the near-circular segment via the analysis of gradient distributions and avoiding arc extraction for all remaining edge segments.

\section{Conclusions}
\label{sec:Conclusions}

Eye tracking is a research topic which spans a large scope including psychology, human-computer interaction, marketing, usability and assistive systems.
And it has recently started to be utilized in VR and AR technologies to provide a more realistic immersion effect.
Pupil detection is an indispensable step in many of these eye tracking applications and have to be performed fast and precisely.
Precision of the algorithm becomes more apparent in applications like VR, AR where the user controls or interacts with other objects and interfaces.
A loose algorithm may cause jitter and significantly effects the quality of experience.

In most studies, pupil detection is handled with straightforward methods which lack accuracy and fail in occlusive cases.
In this study we focused on developing an efficient feature-based algorithm for pupil boundary detection by using the entropy of edge segments.
We basically find elliptical arcs in an input image and try to obtain a final ellipse encircling the pupil with the consensus of all obtained features.

Because the edge segment detection method we employed provide optimum localization, elliptical arcs we extract from the edge segments accurately encircle pupil boundary and estimates its center.
Moreover, by means of the gradient distribution analysis, we boost the execution of the algorithm in an adaptive manner and pave the way for real-time applications for high resolution images.

To make a quantitative assessment, we prepared a comprehensive dataset which consists of 3904 high resolution image collected from 12 subjects.
We performed a comprehensive set of experiments and compared the proposed method with three state of the art algorithms and provided both quantitative and qualitative results.
Experimental evaluations show that APPD algorithm can detect the pupil even in tough occlusive cases without compromising the real-time applicability constraints.
Due to its speed and accuracy, APPD can run in less amount of hardware and therefore it can pave the way for more pervasive eye tracking applications. 


\section*{References}

\bibliography{pupilDetection_refs}

\end{document}